\documentclass[twoside]{article}

\usepackage[accepted]{aistats2023}

\usepackage[round]{natbib}

\usepackage[utf8]{inputenc} 
\usepackage[T1]{fontenc}    
\usepackage{hyperref}       
\usepackage{url}            
\usepackage{booktabs}       
\usepackage{amsfonts}       
\usepackage{nicefrac}       
\usepackage{microtype}      
\usepackage{xcolor}         
\usepackage{amsmath}
\usepackage{graphicx,caption,subcaption}
\usepackage{bm}
\usepackage{array}
\usepackage{wrapfig}
\usepackage{multirow}

\def\x{\mathbf{x}}
\def\xi{\x_{i}}
\def\X{\mathbf{X}}
\def\y{\mathbf{y}}
\def\yi{\y_{i}}
\def\yipred{\tilde{\y}_{i}}
\def\Y{\mathbf{Y}}
\def\s{\mathbf{s}}
\def\w{\mathbf{w}}

\def\E{\mathbb{E}}
\def\sobs{\mathbf{s}_\mathrm{0}}

\def\nnparams{{\bm{\theta}}}

\def\L{\mathcal{L}}
\def\D{\mathcal{D}}

\def\DP{\mathcal{DP}}
\def\Dir{\mathrm{Dir}}
\def\Cat{\mathrm{Cat}}

\def\Normal{\mathcal{N}}
\def\0{\bm{0}}

\def\kl{\mathrm{KL}}

\begin{document}

\twocolumn[
\aistatstitle{Incorporating functional summary information in Bayesian neural networks using a Dirichlet process likelihood approach}
\aistatsauthor{Vishnu Raj \And Tianyu Cui \And Markus Heinonen \And Pekka Marttinen}
\aistatsaddress{Department of Computer Science\\ Aalto University\\ Finland}
]

\begin{abstract}
  Bayesian neural networks (BNNs) can account for both aleatoric and epistemic uncertainty. However, in BNNs the priors are often specified over the weights which rarely reflects true prior knowledge in large and complex neural network architectures. We present a simple approach to incorporate prior knowledge in BNNs based on external summary information about the predicted classification probabilities for a given dataset. The available summary information is incorporated as augmented data and modeled with a Dirichlet process, and we derive the corresponding \emph{Summary Evidence Lower BOund}. The approach is founded on Bayesian principles, and all hyperparameters have a proper probabilistic interpretation. We show how the method can inform the model about task difficulty and class imbalance. Extensive experiments show that, with negligible computational overhead, our method parallels and in many cases outperforms popular alternatives in accuracy, uncertainty calibration, and robustness against corruptions with both balanced and imbalanced data.
\end{abstract}

  \section{Introduction}
Modern deep learning has opened up a plethora of possibilities that previously seemed impossible. Leveraging function approximation capabilities of neural networks, modern deep learning can tackle challenging problems \citep{esteva2019guide,george2018deep,silver2016mastering}, but the black-box nature of neural networks hinders researchers from developing insights into the model's predictions, and the issue is amplified in settings where uncertainty quantification is required. On the other hand, model uncertainty should be calibrated in critical areas such as healthcare and autonomous driving. Bayesian modeling enables a coherent probabilistic perspective for machine learning  \citep{murphy2012machine} and provides valuable tools for data analysis \citep{gelman2014bayesian}. Bayesian neural networks (BNNs) offer a formal framework with promises of improved predictions, reliable uncertainty estimates, principled model comparison, etc \citep{wilson2020case,wilson2020bayesian}.

\begin{figure*}[!t]
    \centering
    \begin{subfigure}[b]{0.22\textwidth}
        \centering
        \includegraphics[width=\textwidth]{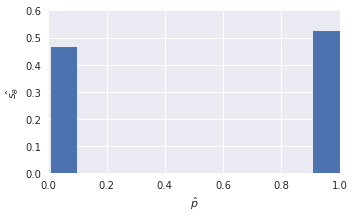}
        \includegraphics[width=0.80\textwidth]{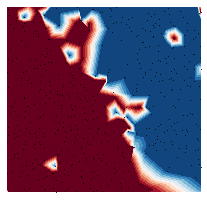}
        \caption{Regular BNN}
        \label{fig:abl-term-1-3}
    \end{subfigure}
    \begin{subfigure}[b]{0.22\textwidth}
        \centering
        \includegraphics[width=\textwidth]{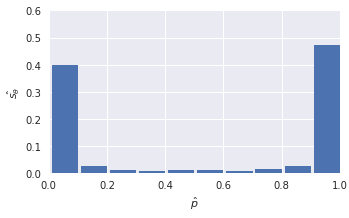}
        \includegraphics[width=0.80\textwidth]{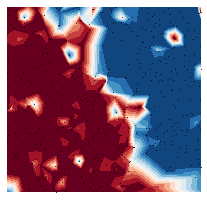}
        \caption{$\sobs = \text{Beta}(0.1, 0.1)$}
        \label{fig:abl-all-term-beta0101}
    \end{subfigure}
    \begin{subfigure}[b]{0.22\textwidth}
        \centering
        \includegraphics[width=\textwidth]{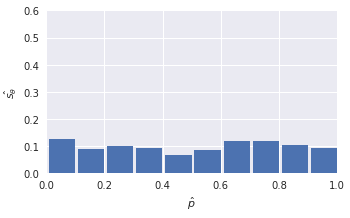}
        \includegraphics[width=0.80\textwidth]{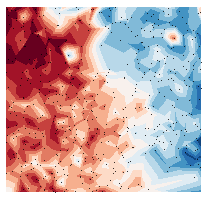}
        \caption{$\sobs = \text{Beta}(1.0, 1.0)$}
        \label{fig:abl-all-term-beta1010}
    \end{subfigure}
    \begin{subfigure}[b]{0.22\textwidth}
        \centering
        \includegraphics[width=\textwidth]{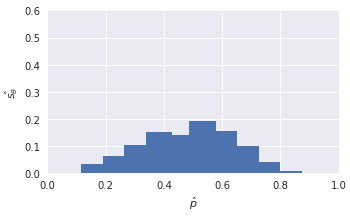}
        \includegraphics[width=0.80\textwidth]{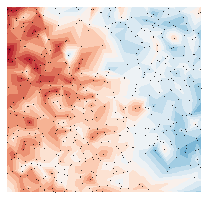}
        \caption{$\sobs = \text{Beta}(5.0, 5.0)$}
        \label{fig:abl-all-term-beta5050}
    \end{subfigure}
    \begin{subfigure}[b]{0.05\textwidth}
        \centering
        \includegraphics[width=0.8\textwidth]{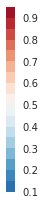}
        \vspace{0.5cm}
        \label{fig:decision-heatmap}
    \end{subfigure}
    \caption{Posterior distribution of predicted sigmoid scores with different summary statistic observations $\sobs$ in binary classification in MNIST. Top row: Posterior distribution of predicted sigmoid samples. Bottom row: Decision surface of the model with samples projected in 2D using t-SNE and colored corresponding to the predicted sigmoid scores. We see from Fig. \ref{fig:abl-term-1-3} that the regular BNN predicts scores peaked towards $0$ or $1$, indicating possible overconfidence. By introducing a likelihood term for the summary statistic the proposed \emph{Summary ELBO} is able to control how the predicted sigmoid scores are distributed, and Figs. \ref{fig:abl-all-term-beta0101} - \ref {fig:abl-all-term-beta5050} show that different $\sobs$ can yield different predicted sigmoid score histograms. This is also evident in the decision surface; the regular BNN has a sharp decision boundary with extreme predicted values while the \emph{Summary ELBO} yields a smoother decision surface.}
    \label{fig:abl-effect-of-sobs}
\end{figure*}

While many works in Bayesian neural networks focus on specifying priors over model parameters \citep{graves2011practical,blundell2015weight} and functional outputs \citep{flam2017mapping,tran2020all,sun2019functional}, there is a surprising gap in incorporating prior knowledge about \textit{summary statistics} of the functional outputs. Such a prior could help in improving uncertainty quantification and calibration of Bayesian neural networks. 
Calibration of neural network predictions is a widely studied topic \citep{guo2017calibration, minderer2021revisiting, wang2021rethinking} and methods such as posterior tempering \citep{wenzel2020good} have been developed. However, these approaches typically deviate from the strictly Bayesian approach by modifying the prior or likelihood with additional parameters. 
Consequently, we study how to incorporate summary statistics information available about a classification task in a fully Bayesian manner. We introduce a formulation where the shape of the distribution of the predicted probabilities (such as sigmoid/softmax scores) is available as prior knowledge, and we demonstrate how such a summary can be informative, e.g., about the difficulty of classification or class imbalance (See Fig. \ref{fig:abl-effect-of-sobs}). Technically, we augment the observed data with this summary, expand the likelihood with a Dirichlet process term for the summary, and derive a formal ELBO for variational training. Through empirical evaluation in multiple classification tasks, we show the proposed approach is able to improve the calibration, robustness and uncertainty of BNNs while maintaining their accuracy. 

The main contributions of this work are,
\begin{enumerate}
    \item We propose a fully Bayesian approach to incorporate summary information into Bayesian neural networks
    \item We introduce how different summary information such as confidence in predictions or class imbalance can be incorporated during model training using the augmented likelihood.
    \item Through comprehensive empirical studies in computer vision and natural language processing, we show that the additional knowledge can in most cases significantly improve the performance of BNNs, especially with corrupted test data or imbalanced classes.
\end{enumerate}

  \section{Background} \label{sec:background}
\subsection{Bayesian neural networks and variational inference}
Different from deterministic neural networks (NNs), Bayesian neural networks (BNNs)
\citep{mackay1992practical,neal2012bayesian} are commonly defined by placing a prior distribution $p(\nnparams)$
on the weights $\nnparams$ of a NN. Moreover, instead of only finding point estimates for
weights $\nnparams$, a posterior distribution of the weights is computed conditionally on the
data according to the Bayes' theorem. Specifically, given a dataset $\D=\{\X, \Y\}$
with inputs $\X=\{\x_{1},\ldots,\x_{N}\}$ and outputs $\Y=\{\y_{1},\ldots,\y_{N}\}$,
we have the likelihood $p(\Y|\X,\nnparams)=p(\Y|f(\X;\nnparams))$ of a BNN on the dataset where
$f(\X;\nnparams)$ is the prediction of the BNN parameterized by $\nnparams$. Then, training a BNN
means computing the posterior distribution $p(\nnparams|\X,\Y)= p(\Y|f(\X;\nnparams))p(\nnparams)/p(\Y|\X)$,
and we predict a new data point $(\x_{\star},\y_{\star})$ by marginalizing out $\nnparams$
from the likelihood according to its posterior 
$p(\y_{\star}|\x_{\star}, \X,\Y)=\int p(\y_{\star}|f(\x_{\star};\nnparams))p(\nnparams|\X,\Y)d\nnparams$.
Unfortunately, neither the posterior of weights nor the predictive distribution of
the new data is analytically tractable for BNNs.

Variational inference can be used to approximate the intractable $p(\nnparams|\X,\Y)$ with
a simpler distribution, $q_{\phi}(\nnparams)$, by minimizing $\text{KL}(q_{\phi}(\nnparams)||p(\nnparams|\X,\Y))$.
This is equivalent to maximizing the Evidence Lower BOund (ELBO) \citep{bishop2006pattern}
\begin{equation}
    \begin{split}
        \mathcal{L}(\phi)=\mathrm{E}_{q_{\phi}(\nnparams)}[\log p(\Y|\X,\nnparams)]-\kl[q_{\phi}(\nnparams)|p(\nnparams)],
        \label{eq: elbo}
    \end{split}
\end{equation}
where the first term is the expected log-likelihood and the second term measures
the divergence between the posterior and the prior. ELBO and its gradients with respect to
$\phi$ can be computed by backpropagation with the reparametrization trick 
\citep{kingma2013auto}. Therefore, the posterior predictive distribution can be 
approximated by  
\begin{align}
        p(\y_{\star}|\x_{\star}, \X,\Y) 
            & \approx \int p(\y_{\star}|f(\x_{\star};\nnparams))q_{\phi}(\w)d\nnparams \nonumber \\
            & \approx \frac{1}{M}\sum_{l=1}^{M}p(\y_{\star}|f(\x_{\star};\nnparams^{(l)})),
\end{align}
where $\nnparams^{l} \sim q_{\phi}(\nnparams)$ and $M$ is the number of Monte Carlo
samples drawn from posterior distribution.

\begin{figure*}[!t]
    \centering
    \begin{subfigure}[b]{0.27\textwidth}
        \centering
        \includegraphics[width=\textwidth]{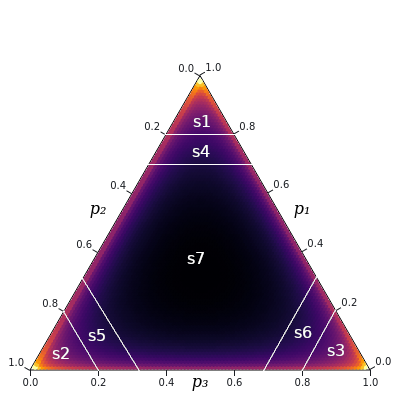}
        \caption{Easy task, balanced data}
        \label{fig:dd-clean}
    \end{subfigure}
    \begin{subfigure}[b]{0.27\textwidth}
        \centering
        \includegraphics[width=\textwidth]{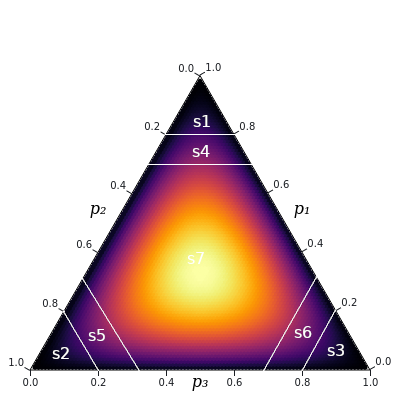}
        \caption{Difficult task, balanced data}
        \label{fig:dd-noisy}
    \end{subfigure}
    \begin{subfigure}[b]{0.27\textwidth}
        \centering
        \includegraphics[width=\textwidth]{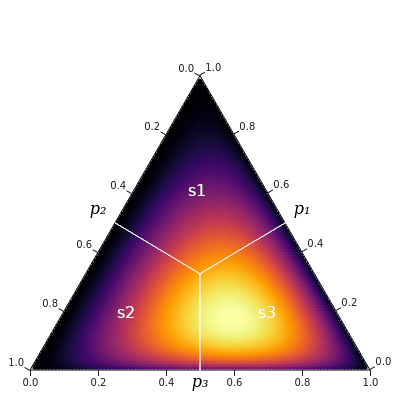}
        \caption{Imbalanced data}
        \label{fig:dd-noisy-im}
    \end{subfigure}
    \caption{Different examples for selecting $\sobs$ to reflect prior information in a $3$ class classification setting. Figure shows the heatmap of the distribution of sigmoid scores where dark means low probability and bright means high probability. Here, $\sobs$ is a Dirichlet distribution and, by selecting the parameters of the Dirichlet distribution appropriately, our method provides a flexible and principled approach to incorporate prior information on the difficulty of the classification task and class imbalance. The figure also shows the binning of the simplex into a finite number of regions required by the finite approximation in DP inference. See text for details.}
    \label{fig:dd-vis}
\end{figure*}

\subsection{Dirichlet processes} \label{sub:dp}
The Dirichlet process (DP) \citep{teh2010dirichlet} is a stochastic process widely used in Bayesian nonparametrics. Different from Gaussian processes (GPs) \citep{seeger2004gaussian}, which model distributions over functions with Gaussian marginals, DPs are stochastic processes over probability measures with Dirichlet marginals. In machine learning, DPs have been used as an infinite-dimensional generalization of the Dirichlet distribution in mixture models \citep{neal1992bayesian} and in topic modeling \citep {teh2006hierarchical}. A DP, $G\sim\DP(H,\alpha)$, is parameterized by the \textit{base measure} $H$, which is a distribution $H$ over a probability space $\Theta$, and the \textit{concentration parameter} $\alpha$, a positive real number, such that 
\begin{equation}
    \begin{split}
        (G(A_1),\ldots,G(A_b))\sim \Dir(\alpha H(A_1), \ldots, \alpha H(A_b))
    \end{split}
    \label{def:dp-finite}
\end{equation}
for every finite measurable partition $\{A_1,\ldots,A_b\}$ of $\Theta$. The base measure $H$ is the mean of the DP, i.e., for any $A\subset\Theta$, $\E[G(A)]=H(A)$, and it specifies the overall shape of $G$. The concentration parameter $\alpha$ serves as the inverse variance of the DP \citep{teh2010dirichlet}, such that a large $\alpha$ will force $G$ to be close to $H$, see Fig. \ref{appfig:dp-alpha} in Appendix for examples of sampled distributions from a DP with different $\alpha$.

  \section{Incorporating summary information} \label{sec:proposed}
We consider multiclass classification using a Bayesian neural network, where target $\yi$ is encoded label, such as the one hot encoding. Let $\yipred = f(\xi; \nnparams)$ be the prediction by the neural network for input $\xi$. In binary classification, we have $\yipred \in [0, 1]$ representing the probability of one of the classes. In multiclass classification, $\yipred \in [0, 1]^K$ with $\sum_{k=1}^{K} \tilde{y}_{ik} = 1$, where $\tilde{y}_{ik}$ is the probability of class $k$ and $K$ is the number of classes.

Ideally, $\yipred$ would be equal to $\yi$, corresponding to the perfect prediction. However, in practice in multi-class classification we get $\yipred \in [0, 1]^K$, where each entry in the predicted vector $\yipred$ is the normalized score of a particular class, corresponding to the probability that the sample belongs to the class. Here, we assume that the modeler has access to a summary statistic $\sobs$ representing the how the predicted probabilities $\yipred$ are distributed over the dataset. For example, in binary classification the summary statistic $\sobs$ is a distribution in the range $[0, 1]$ (e.g. a Beta distribution, Fig. \ref{fig:abl-effect-of-sobs}) and in multiclass classification $\sobs$ is a distribution over the prediction simplex (e.g. a Dirichlet, Fig. \ref{fig:dd-vis}).

The summary statistic $\sobs$ is then used for controlling the distribution of predicted sigmoid/softmax scores $\yipred$ according to available prior knowledge. In practice this prior knowledge should reflect considerations external to the current data set. For example, if we know that the classification task is easy and well separable, 
we expect $\yipred$ to take values where one entry is close to $1$ while others are close to $0$, 
which we can represent with a summary statistic $\sobs=\text{Dir}(\alpha_1,\alpha_2,\ldots,\alpha_k)$,
with all parameters $\alpha_k < 1$ (see Fig. \ref{fig:dd-clean}). Conversely, if we know that the dataset is noisy and/or not easily separable, we would expect the predictions $\yipred$ to concentrate towards the center of the prediction simplex, which could be represented with a Dirichlet with $\alpha_k > 1$ (Fig. \ref{fig:dd-noisy}). Furthermore, the relative magnitudes of the different $\alpha_k$ parameters can inform about the frequencies of different labels in imbalanced data (Fig. \ref{fig:dd-noisy-im}).

\begin{figure}[!h]
    \centering
    \vspace{-3mm}
    \begin{subfigure}[b]{0.24\textwidth}
        \centering
        \includegraphics[width=\textwidth]{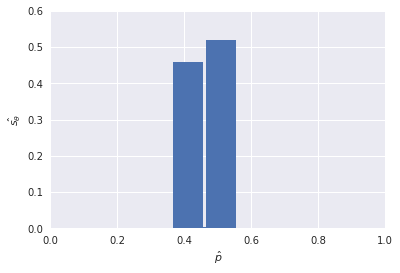}        
        \caption{Independent Dirichlet}
        \label{fig:modeling-dl.png}
    \end{subfigure}\hfill%
    \begin{subfigure}[b]{0.24\textwidth}
        \centering
        \includegraphics[width=\textwidth]{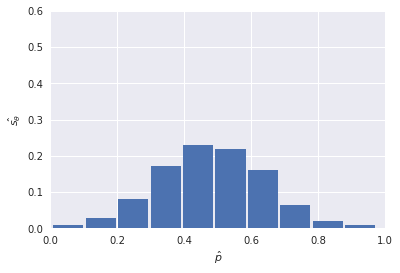}
        \caption{DP modeling}
        \label{fig:modeling-dp.png}
    \end{subfigure}
    \caption{We show the posterior distribution of $\yipred$s when using $\sobs=\text{Beta}(5.0, 5.0)$ with different approaches. Modeling sigmoid outputs independently can result in all predictions concentrating near the mode of the distribution as shown in Fig. \ref{fig:modeling-dl.png}. However, the proposed DP modeling avoids this and tries instead to match the whole distribution of the scores set by $\sobs$ as shown in Fig. \ref{fig:modeling-dp.png}.}
    \label{fig:why-dp-model}
    \vspace{-6mm}
\end{figure}

As another point to emphasize, our approach considers how $\yipred$ are collectively (i.e. jointly) distributed in the prediction simplex for all samples. Another option would be to model $\yipred$ with a Dirichlet separately for each $i$, similar to \cite{sensoy2018evidential}. However, this would make all $\yipred$s concentrate near the mode of the Dirichlet (see Fig. \ref{fig:why-dp-model}). Instead, we want to control how $\yipred$s are distributed across all samples from training dataset $\D$, indicating that the dataset will contain both easy and difficult samples to classify.

\subsection{Incorporating prior knowledge through sequential Bayesian inference} \label{sub:informative-prior}
We want to train a Bayesian neural network $f(\x ; \nnparams)$ for classification from a dataset $\D = (\X, \Y) = \{\xi,\yi\}_{i=1}^N$ of $N$ observations from the input space $\xi \in \mathcal{X}$ and output space $\yi \in \mathcal{Y}$ with $K$ labels, where $\nnparams$ denote the parameters of the neural network. We also consider a summary statistic $\s_{\nnparams}=S(\nnparams,\X) = S(f(\x_1; \nnparams), \ldots,f(\x_N; \nnparams))$, where the function $S$ calculates the distribution of the \emph{predicted} sigmoid scores $f(\x_i; \nnparams)$ in the training set. This can be a continuous or discrete density estimator. In addition, we denote with $\sobs$ the \emph{observed} summary statistic, which corresponds to the distribution of sigmoid/softmax scores available from prior knowledge, representing information about label distribution (in case of class imbalance), or mass in different parts of the prediction simplex  (in case of difficult classification tasks).

\begin{figure}[!h]
    \centering
    \begin{subfigure}[b]{0.105\textwidth}
        \centering
        \includegraphics[width=\textwidth]{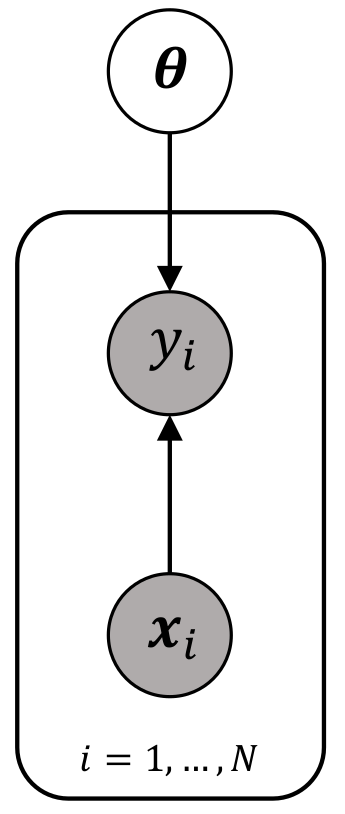}
        \caption{BNN}
        \label{fig:gm-elbo.png}
    \end{subfigure}\hspace{1cm}%
    \begin{subfigure}[b]{0.175\textwidth}
        \centering
        \includegraphics[width=\textwidth]{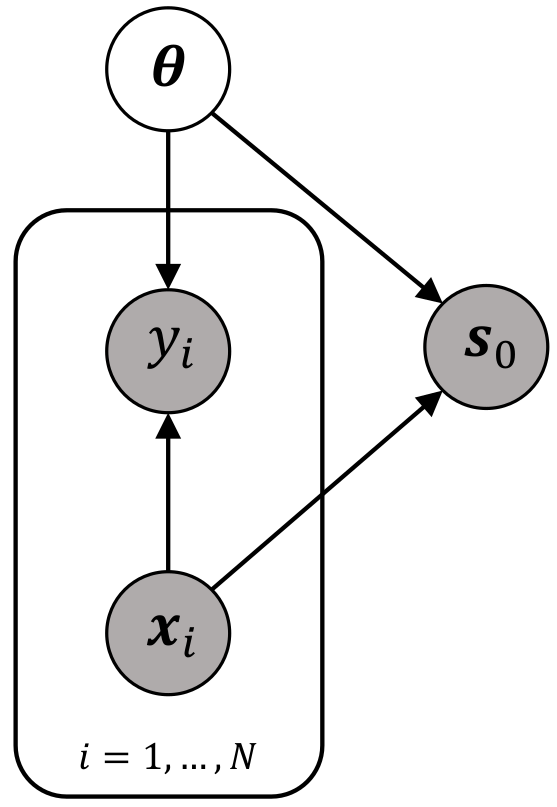}
        \caption{BNN with summary}
        \label{fig:gm-selbo.png}
    \end{subfigure}
    \caption{Graphical model for BNNs. Fig. \ref{fig:gm-elbo.png} shows the graphical model for vanilla BNN where model parameters $\bm{\theta}$
        are only related to label $\y$. In our proposed summary likelihood model Fig. \ref{fig:gm-selbo.png}, we show how to model additional summary
        statistics information. We model the summary information $\sobs$ as derived from input variable $\x$ and model parameters
        $\nnparams$, and is an observed node in the model.}
    \label{fig:gm}
\end{figure}

Assume now that we have a prior $p(\nnparams)$ for the neural network weights, and we observe an augmented data $\D_{\text{aug}} = (\D,\sobs)$ where $\sobs$ is the observed summary statistic and $\D = (\X, \Y)$ are the observations.  Fig. \ref{fig:gm} shows a graphical model assumed by a traditional BNN and compares that with our joint model with summary information $\sobs$. Specifically, we assume that the joint distribution factorizes as follows:
\begin{equation}
    p(\X, \Y, \sobs, \nnparams) 
        = \left[ \prod_{i=1}^N p(\y_i | \x_i, \nnparams) \right] 
                p(\sobs | \X, \nnparams) p(\nnparams). \label{eqn:sobs-model}
\end{equation}
Hence, the approach can be formally seen as sequential Bayesian inference, which first updates the non-informative prior $p(\nnparams)$ into an informative prior by multiplying with the \textit{summary likelihood}, $p(\sobs | \X, \nnparams)$, and then uses the informative prior for modeling the data $(\X,\Y)$ using the regular likelihood. We define the summary likelihood as 
\begin{equation}
    p(\sobs | \X, \nnparams) = \DP(\sobs | \s_{\nnparams}, \alpha).
\end{equation}
In other words, the observed summary $\sobs$ is distributed as a Dirichlet process whose base measure is equal to $\s_\nnparams=S(\nnparams,\X)$, i.e., the histogram of sigmoid/softmax outputs predicted by the NN for the training data, and a concentration hyperparameter $\alpha$. Consequently, with a large $\alpha$ the predicted and observed summary statistics $\s_\nnparams$ and $\sobs$ are expected be close to each other. The model definition is completed by defining the likelihood as a categorical distribution:
\begin{equation}
p(\y_i | \x_i, \nnparams)=\Cat\big(\y_i | f(\x_i, \nnparams) \big), \text{for all } i,
\end{equation}
and the prior conventionally as $p(\nnparams)=N_\nnparams(\bm{\mu},\bm{\Sigma})$.

One way to think about the summary statistic, $\sobs$, is to interpret it as a `pseudo-observation'; pseudo-observations are often used to interpret common priors \citep{gelman2014bayesian}. To understand why the prior knowledge is incorporated through another likelihood term $p(s_0|X_{obs},\theta)$, it is instructive to notice that a prior on the parameters $\theta$, $p(\theta)$, already induces a prior on the distribution of outputs, $s_{\theta}$, which we can here denote  by $p_\theta(s_\theta)$. In general, when there exists some prior knowledge, captured by $s_0$, about $s_\theta$, an obvious thing would be to define a prior distribution, something like $p(s_\theta|s_0)$, or equivalently a joint distribution $p(s_\theta,s_0)$. The problem is that there can't be two prior distributions: $p_\theta(s_\theta)$ and $p(s_\theta|s_0)$, for the same quantity $s_\theta$. Instead, we calculate $p(s_\theta|s_0)$ according to the formal Bayesian procedure where we update the initial prior distribution $p_\theta(s_\theta)$ into $p(s_\theta|s_0)$ using the Bayes' rule, which happens through the multiplication of the previous prior using a likelihood term. Consequently, as we show later, this yields a well-defined ELBO corresponding to proper Bayesian inference. Further, \cite{gelman2021stan} suggests to consider one prior as data when there are two sources of prior knowledge for the same parameter instead of two priors, because the former is more consistent with Bayesian theory.

\subsection{Inference with summary ELBO} \label{sub:selbo}
In this setting our goal is simply to infer the parameter posterior $p(\nnparams | \D)$, which we approximate variationally with $q_\phi(\nnparams)$. This induces an ELBO
\begin{align}
    \L(\phi) 
        &= \E_{q_\phi(\nnparams)} \log p(\D|\nnparams) - \kl[ q_\phi(\nnparams) || p(\nnparams) ] \nonumber \\
        &= \E_{q_\phi(\nnparams)} \left[ \sum_{i=1}^{N} \log p(\y_i | \x_i, \nnparams) 
                    + \log p(\sobs | \X, \nnparams) \right] \nonumber \\
                & \qquad \qquad - \kl[ q_\phi(\nnparams) || p(\nnparams) ] \nonumber \\
        &= \E_{q_\phi(\nnparams)} \left[ 
                    \sum_{i=1}^N \log \Cat\big(\y_i | f(\x_i, \nnparams) \big) \right. \nonumber \\
                    & \left. \qquad \qquad
                        + \log \DP\big(\sobs | \s_\nnparams, \alpha\big) 
                    \right] \nonumber \\
                    & \qquad \qquad -\kl \Big[ 
                        \Normal_\nnparams(\bm{\mu}_\phi,\bm{\Sigma}_\phi) \, || \, \Normal_\nnparams(\0, \sigma^2 I) 
                    \Big] \nonumber \\
        &\approx \frac{1}{M} \sum_{i=1}^{N}\sum_{j=1}^M \log \Cat\big(\y_i | f(\x_i, \nnparams_j) \big) \nonumber \\
                & \qquad \qquad + \frac{1}{M} \sum_{j=1}^M \log \DP\big(\sobs | \s_{\nnparams_j}, \alpha\big) \nonumber \\
                & \qquad \qquad - \kl\Big[ 
                        \Normal_\nnparams(\bm{\mu}_\phi,\bm{\Sigma}_\phi) \, || \, \Normal_\nnparams(\0, \sigma^2 I)
                    \Big]   \label{eqn:summary-elbo},
\end{align}
where $\{\nnparams_j\}_{j=1}^M \sim q_\phi(\nnparams)$ are the samples from the inferred posterior and $M$ is the number of Monte Carlo samples. Hence, compared to the traditional ELBO \citep{kingma2013auto,neal1992bayesian}, our objective,  \emph{Summary ELBO}, defined in (\ref{eqn:summary-elbo}), incorporates prior information about the modeler's belief on how the predictions $\yipred$ should be jointly distributed, as captured by the observed summary statistic $\s_0$.

\subsection{Computation of the DP summary likelihood term}
One challenge in the objective function (\ref{eqn:summary-elbo}) is the computation of summary likelihood involving the DP (second term). As a closed form expression is unavailable \citep{teh2010dirichlet}, we use a finite partition approximation, where the parameter space $\Theta$ is divided into a finite number of bins $\{A_1,\ldots,A_b\}$ corresponding to discretized histograms and the likelihood is evaluated using Eqn. (\ref{def:dp-finite}). In the binary experiments, the softmax scores are distributed in the $[0,1]$ interval, which we divide into multiple bins, as demonstrated in Fig.\ref{fig:abl-term-1-3}-\ref{fig:abl-all-term-beta5050}. In the multiclass classification we divide the prediction simplex symmetrically into regions where some regions are more central and some in the corners of the simplex, allowing us to express prior knowledge about task difficulty, i.e., how much of the probability mass should be given to uncertain predictions corresponding to sigmoid scores near 0.5 and how much to confident predictions with scores close to 1, demonstrated in Figs \ref{fig:dd-clean} and \ref{fig:dd-noisy}. In the imbalanced data experiment, we use a partition shown in Fig. \ref{fig:dd-noisy-im}, which accounts for the total mass allocated to each predicted class, but is agnostic about how far the score is from the center of the simplex.

In practice, the predictions $\yipred$ from the model at each training step are collected from the entire minibatch and a histogram over the specified regions is constructed. To be able to backpropagate through the operation, we use SoftHistogram to construct the histogram in our experiments. The SoftHistogram function identifies the total mass in each bin using a pair of sigmoid functions and aggregating over the minibatch. Details of SoftHistogram construction is discussed in Appendix \ref{appsec:softhist}. The major bottleneck in using this approximation is the quality of SoftHistogram results. To address this, we use a moderately large minibatch size to make the estimation less noisy.

  \section{Related works} \label{sec:related-works}
\textbf{Functional BNNs priors} Our approach can be seen as a way to incorporate summary information about the predictive distribution, and hence it is conceptually related to functional priors. Previously, Gaussian processes have been proposed to encode rich functional structures as prior knowledge. \cite{flam2017mapping} and \cite{tran2020all} transformed a functional GP prior into a weight-space BNN prior by minimizing the Kullback–Leibler divergence and Wasserstein distance respectively. Functional BNNs \citep{sun2019functional} performed variational inference directly with GP priors. Other recent works which concern with the output behavior include Noise contrastive priors (NCPs) \citep{hafner2018noise} and Output-Constrained BNNs \cite{yang2020incorporating}. A comprehensive review of deep learning priors is given in \cite{fortuin2021priors}.

\textbf{Weight-space BNN priors} 
In the weight space, a fully factorized Gaussian prior has been proposed by \cite{graves2011practical} and \cite{blundell2015weight}, and interpreted as equivalent to dropout when using a mixture of Dirac-deltas as the variational posterior \citep{gal2016dropout}. \cite{nalisnick2019dropout} extended these works and interpreted NNs with any multiplicative noise as BNNs with a Gaussian scale mixture prior \citep{andrews1974scale} and Automatic Relevance Determination (ARD) \citep{mackay1994bayesian}. Moreover, low-rank priors, such as the k-tied normal \citep{swiatkowski2020k} and rank-1 perturbation \citep{dusenberry2020efficient}, were combined with ensemble methods \cite{lakshminarayanan2017simple} to capture multiple modes, and they had better convergence rates. To model the correlation between the weights, Matrix-variate Gaussian priors were proposed by \cite{neklyudov2017structured} and \cite{sun2017learning}. Also sparse priors have been defined, such as the log-uniform \citep{molchanov2017variational, louizos2017bayesian}, log-normal \citep{neklyudov2017structured}, horseshoe \citep{louizos2017bayesian, ghosh2018structured}, and spike-and-slab priors \citep{deng2019adaptive}. \cite{cui2021informative} proposed a two-stage procedure to encode the prior knowledge about the data signal-to-noise ratio into a Gaussian scale mixture prior. Overall, it is often challenging to incorporate more general domain knowledge other than sparsity into the weight-space priors. 

\textbf{Evidential Deep Learning}
Different from ordinary deep learning, which is trained to predict the parameters of the likelihood function with Maximum Likelihood, evidential deep learning (EDL) is trained to predict the parameters of likelihood with the Type II Maximum Likelihood (ML-II, i.e., maximizing the model evidence). Therefore, the model predictions, as well as the aleatoric and epistemic uncertainty estimations, come from the learned prior of the likelihood. In the classification setting, \cite{sensoy2018evidential,malinin2018predictive} proposed to learn a Dirichlet prior of the categorical likelihood parameters, and in regression, \cite{amini2020deep} learned a Normal Inverse-Gamma prior of the Gaussian likelihood. 
Although EDL provides a reasonable uncertainty estimation, a heuristic regularization on evidence has to be applied to avoid over-fitting due to the ML-II. 

\textbf{Non-Bayesian approaches of incorporating domain knowledge} Sophisticated regularization techniques, i.e., explanation prior, have been proposed for deterministic NNs to incorporate extra domain knowledge \citep{ross2017right}. When the importance score of each feature is known \textit{a priori}, attribution priors were proposed to regularize the feature importance of the model to agree with the prior importance score, such as DeepSHAP \citep{tseng2020fourier} and Contextural Decomposition \citep{rieger2020interpretations}. DAPr \citep{weinberger2020learning} matched the feature attribution to a learned prior feature importance from meta-features. In a genetics application, MEP \cite{cui2021improving} was proposed to incorporate the feature main effects (i.e., linear regression coefficients) on an external large dataset into NNs on a small dataset.
  \section{Experimental results}  \label{sec:results}
Here, we show the utility of the proposed method, abbreviated as \textbf{S-ELBO} for \textit{Summary ELBO}, in classification tasks from computer vision and natural language processing domains. Specifically, we show results for image classification and sentiment analysis. The experiments cover both binary and multiclass classification tasks. In all the cases, Mean Field Variational Inference (MFVI) with $\Normal(\0, \sigma_0^2 \bm{I})$ prior is used to train the neural networks. We cross-validate the prior variance $\sigma_0 \in \{0.10, 0.25, 0.50, 1.00, 2.00\}$, but as the results are not sensitive to this choice (Fig. \ref{fig:cv-sigma0} in Appendix), we use the default value $\sigma_0=1$ throughout. For the numerical results, each model is trained $5$ times independently, and the mean and standard error of each metric are reported. The code is implemented in PyTorch \cite{paszke2019pytorch} and available at \href{https://github.com/v-i-s-h/summary-likelihood}{\texttt{github.com/v-i-s-h/summary-likelihood}}. 

For modeling the Dirichlet Process likelihood term, we cross validate the concentration parameter $\alpha \in \{10, 50, 100, 500, 1000, 2500, 5000, 10000\}$ on a separate validation set. For the summary prior histogram $s_0$ in the DP likelihood, we cross validate between the uniform distribution and a distribution based on automatic parameter selection as described in Appendix \ref {appsec:derivation-prior-params}. Models are trained for $3000$ steps for binary and $5000$ steps for multiclass classification problems. We use minibatch size of $256$ and Adam optimizer with a constant learning rate $10^{-3}$. We report the negative log-likelihood (NLL), accuracy, and Expected Calibration Error (ECE) \citep{guo2017calibration} in clean (in-domain) and corrupted test sets. In the detection of out-of-distribution samples we report the difference in predictive entropy $\Delta_\text{OOD}$ between OOD and in-domain samples, and in the multi-class classification with imbalanced classes we report the F1 score. All scores are reported for held-out test sets. During cross validation, the optimal hyperparameters are chosen based on the NLL. We compare against the vanilla BNN trained with the standard ELBO (\textbf{ELBO}) \citep{blundell2015weight}  , Evidential Deep Learning (\textbf{EDL}) \citep{sensoy2018evidential} and Label Smoothing (\textbf{LS}), all sharing the same NN architecture in the same task. ELBO and LS are trained with MFVI, using the same prior as S-ELBO. For LS, we cross validate the smoothing factor $\epsilon \in \{0.01, 0.05, 0.10\}$. For EDL, we use the setup recommended in \cite{sensoy2018evidential}, and train it in multiclass classification with an annealing step of $1000$.

\subsection{Sentiment analysis task} \label{ss:sentiment-analysis}
Sentiment analysis is an NLP task of classifying the polarity of a given text, usually posed as binary classification. However, the analysis of phrases from each of the sample texts \citep{socher2013recursive} shows that the contituent phrases can have intermediate values of sentiment, not fully captured by the binary labels. Hence, sentiment analysis is a perfect example of a  classification task where labels are not always too confident. We use Stanford Sentiment Treebank \cite{socher2013recursive} as our source data with labels and use Sentence-BERT \citep{reimers2019sentence} to compute a $768$ dimensional embedding for each sample text. A feedforward BNN with a single hidden layer of dimension $128$ is trained on these embeddings using the alternative methods. For training the proposed method, a uniformly distributed prior histogram $\s_0$ is assumed, reflecting the inherent uncertainty in the labels, and discretized into $10$ regions of equal width in [0,1], similarly to the examples in Fig. \ref{fig:abl-effect-of-sobs}.

\begin{table*}[!h]
    \caption{Results on multiclass classification task with CIFAR10 dataset. Hyperparameters are cross validated using validation NLL - for the proposed method, we used $\alpha=1000$ and for LS, we used $\epsilon = 0.01$. For OOD experiments, we used SVHN dataset as test data. Detailed results are
    given in Tables \ref{apptab:cifar10-indomain-overview} - \ref{apptab:cifar10-ood-overview} in Appendix \ref{appsec:cifar10-additional-results}.}
    \label{tab:cifar10-overview}
    \centering
    \small
    \begin{tabular}{l c c c c c c c}
        \toprule
               & \multicolumn{3}{c}{In-domain testset} & \multicolumn{3}{c}{Corrupted testset} & OOD testset \\
               \cmidrule(lr){2-4} \cmidrule(lr){5-7}
        Method & NLL$^{\downarrow}$ & Accuracy$^{\uparrow}$ & ECE$^{\downarrow}$ & NLL$^{\downarrow}$ & Accuracy$^{\uparrow}$ & ECE$^{\downarrow}$ & $\Delta_\text{OOD}$$^{\uparrow}$\\
        \midrule
            ELBO     &          $0.76 \pm 0.01$ & $\mathbf{0.82 \pm 0.00}$ &          $0.10 \pm 0.00$&          $1.34 \pm 0.02$ & $\mathbf{0.70 \pm 0.00}$ &          $0.18 \pm 0.00$ &          $0.55 \pm 0.08$ \\
            LS       &          $1.93 \pm 0.02$ &          $0.79 \pm 0.00$ &          $0.17 \pm 0.00$&          $3.36 \pm 0.09$ &          $0.67 \pm 0.01$ &          $0.27 \pm 0.01$ &          $0.13 \pm 0.03$ \\
            EDL      &          $0.78 \pm 0.01$ & $\mathbf{0.82 \pm 0.00}$ & $\mathbf{0.08 \pm 0.00}$&          $1.31 \pm 0.02$ &          $0.68 \pm 0.00$ & $\mathbf{0.16 \pm 0.00}$ & $\mathbf{0.80 \pm 0.08}$ \\
            Proposed & $\mathbf{0.68 \pm 0.01}$ & $\mathbf{0.82 \pm 0.00}$ & $\mathbf{0.08 \pm 0.00}$& $\mathbf{1.23 \pm 0.02}$ & $\mathbf{0.70 \pm 0.00}$ & $\mathbf{0.16 \pm 0.00}$ &          $0.54 \pm 0.03$ \\
        \bottomrule
    \end{tabular}
\end{table*}

\begin{figure*}[!h]
    \centering
    \includegraphics[width=0.72\textwidth]{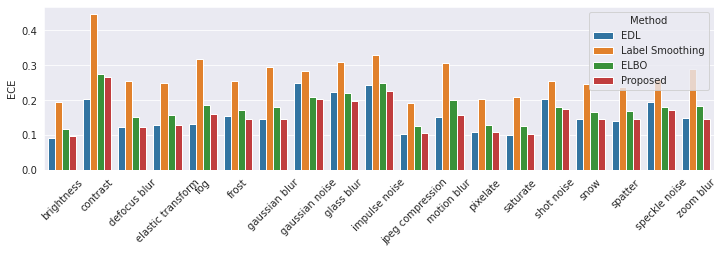}
    \caption{Comparison of ECE on different corruptions. The models are trained on
    clean CIFAR10 data and tested with various corruptions from the CIFAR-10-C dataset.}
    \label{fig:cifar10c-ece-comparison}
\end{figure*}

\begin{table}[!h]
    \caption{Comparison of different methods on the sentiment analysis task. All models achieve $>88\%$ accuracy. Comprehensive comparison of different variants of the proposed method along with accuracy and AUROC metrics is given in Table \ref{apptab:sst-indomain-overview} in appendix. $^{\downarrow}$ means lower the better.}
    \label{tab:sst-indomain-overview}
    \centering
    \small
    \begin{tabular}{l c c c}
        \toprule
        Method & NLL$^{\downarrow}$ & ECE$^{\downarrow}$ \\
        \midrule
            ELBO                    &          $0.341 \pm 0.024$ &          $0.045 \pm 0.009$ \\
            LS($\epsilon = 0.05$)   &          $0.444 \pm 0.029$ &          $0.071 \pm 0.004$ \\
            EDL                     &          $0.301 \pm 0.001$ &          $0.044 \pm 0.004$ \\
            Proposed ($\alpha=1000$)& $\mathbf{0.288 \pm 0.002}$ & $\mathbf{0.026 \pm 0.001}$ \\
        \bottomrule
    \end{tabular}
\end{table}

\textbf{In-domain prediction.} Summary results for an in-domain prediction task (clean test data) are given in Table \ref{tab:sst-indomain-overview}. While all methods achieve $>88\%$ accuracy (Table \ref{apptab:sst-indomain-overview} in Appendix), the proposed method is able to provide significantly better NLL and calibration performance. We explore alternative prior histograms $s_0$ in Appendix (\ref{appsec:sstbert-additional-results}) and show that incorporating the information about uncertainty in constituent phrases helps the model to improve both calibration as well as prediction accuracy.

\begin{figure}[!h]
    \centering
    \includegraphics[width=\linewidth]{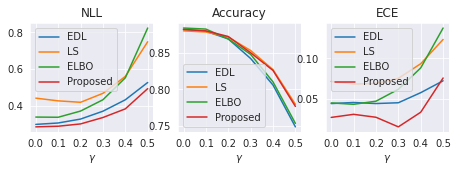}
    \caption{Comparison of different methods in corrupted test embeddings for the sentiment analysis task. $\gamma$ represents the strength of noise added. A detailed comparison is available in Fig. \ref{appfig:sstbert-corruption-comparison} in appendix.}
    \label{fig:sstbert-corruption-comparison}
\end{figure}

\textbf{Corrupted test data.} To study the robustness of the methods against data corruptions, we perturb the BERT embeddings in test data with variance preserving noise as $\tilde{\mathbf{e}} = (1-\gamma) * \mathbf{e} + \gamma * \eta$, where $\mathbf{e}$ is the noise free embedding and $\eta \sim \Normal (\0, \mathbf{I})$. The results in Fig. \ref{fig:sstbert-corruption-comparison}  show that the proposed method is robust against corruptions. Even though both ELBO and the proposed method are trained with MFVI, we see that robustness of ELBO deteriorates significantly with added noise while the proposed method better retains its robustness, even compared to EDL.

\subsection{Multiclass classification with CIFAR10}
Here we consider multiclass classification with CIFAR10 data and balanced classes. For OOD experiments, we use the SVHN dataset \citep{netzer2011reading} and for corruptions, CIFAR-10-C \citep{hendrycks2019robustness}.

\begin{table*}[!h]
    \caption{Results with imbalanced data. Models are trained on Imbalanced CIFAR10. The parameters of each model are selected using the validation NLL. LS is not included due to a very large NLL values with corrupted test data.}
    \label{tab:cifar10im-overview}
    \centering
    \small
    \begin{tabular}{l c c c c}
        \toprule
               & \multicolumn{2}{c}{In-domain test data} & \multicolumn{2}{c}{Corrupted test data} \\
               \cmidrule(lr){2-3} \cmidrule(lr){4-5}
        Method & NLL$^{\downarrow}$ & F1 Score$^{\uparrow}$ & NLL$^{\downarrow}$ & F1 Score$^{\uparrow}$ \\
        \midrule
            ELBO                    &          $1.158 \pm 0.026$ & $\mathbf{0.849 \pm 0.002}$ &          $6.354 \pm 0.117$ &          $0.331 \pm 0.012$ \\
            EDL                     & $\mathbf{0.703 \pm 0.009}$ &          $0.824 \pm 0.003$ & $\mathbf{2.840 \pm 0.034}$ &          $0.314 \pm 0.012$ \\
            Proposed ($\alpha=500$) &          $0.960 \pm 0.022$ & $\mathbf{0.847 \pm 0.001}$ &          $3.564 \pm 0.070$ & $\mathbf{0.400 \pm 0.011}$ \\
        \bottomrule
    \end{tabular}
\end{table*}

\begin{figure*}[!h]
    \centering
    \includegraphics[width=0.72\textwidth]{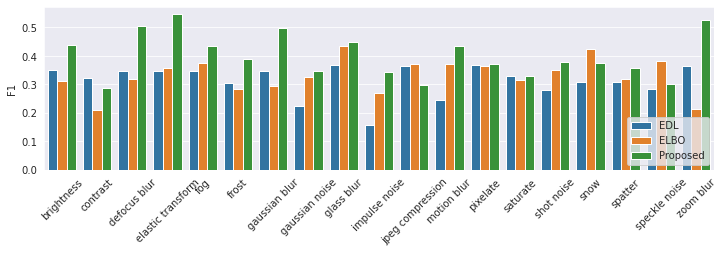}
    \caption{Comparison of F1 on different corruptions on imbalanced dataset. Classifier is trained clean imbalanced CIFAR10 data and tested with various corruptions from CIFAR-10-C dataset.}
    \label{fig:cifar10imc-f1-comparison}
\end{figure*}

\textbf{In-domain prediction.} We give the results on in-domain prediction in Table \ref{tab:cifar10-overview}. While we observe that none of the considered methods reaches the s-o-t-a accuracy for VGG11, we nevertheless clearly see that training the models with the proposed method can reduce the ECE significantly when compared to the regular BNN trained with ELBO, and closely matches EDL using the same architecture. 

\textbf{Corrupted test data.} When testing on different corruptions, we can observe in  Table \ref{tab:cifar10-overview} and Fig. \ref{fig:cifar10c-ece-comparison} that the proposed method singificantly improves NLL and ECE compared with the vanilla ELBO, which does not incorporate the prior summary information. This demonstrates how the summary prior helps the model to regularize its predictions. In ECE the proposed method and EDL are jointly the best, whereas in NLL the proposed method is the single best method with corrupted test data.

\textbf{Detection of out-of-distribution samples.} Here we compare in-distribution and OOD predictive entropies. A model that captures uncertainty properly should have a smaller predictive entropy (larger confidence) for in-distribution than for OOD test samples. Here, EDL performs better with a larger $\Delta_\text{OOD}$, but this comes at the cost of higher in-domain entropy (Table \ref{apptab:cifar10-ood-overview} in Appendix). Instead, incorporating prior summary information through S-ELBO  appropriately balances between in-domain and OOD prediction confidence.

\subsection{CIFAR10 with class imbalance}
To test the ability of the method to incorporate prior knowledge about class imbalance, we create a dataset from CIFAR10 by sub-sampling image classes, such that the imbalance ratio is $1:1/2:1/4:\ldots:1/2^8:1/2^8$. We assume that the class fractions are available as prior knowledge, and use this to construct $\s_0$ as demonstrated in Fig. \ref{fig:dd-noisy-im}.

\textbf{In-domain prediction.} The results of in-domain prediction are given in Table \ref{tab:cifar10im-overview}. Here EDL has a better NLL, but the proposed method and ELBO yield a much larger F1 score. As the NLL is dominated by the majority class, we conclude that the proposed method does a better job in the challenging task of classifying the minority class items. Note that the summary likelihood regularizes the predicted scores to reflect the class imbalance and hence the proposed model is able to provide both a low NLL (relative to ELBO) and a high F1-score.

\textbf{Corrupted test data.} More useful insights can be obtained when analysing the performance under corrupted test data. Table \ref{tab:cifar10im-overview} and Fig. \ref{fig:cifar10imc-f1-comparison} show the F1 scores of different methods for corrupted test datasets. These results show that incorporating the prior knowledge of class imbalance provides a significant improvement in F1 score compared to both ELBO and EDL. This clearly points to the fact that the strategy of the proposed method to allocate probability mass for each class label through the summary likelihood results in robust predictions under noisy input.

\textbf{Additional experiments.} We also provide a comprehensive study of binary classification on a dataset derived from MNIST in Appendix \ref{appsec:bmnist-additional-results}. We study different dataset sizes and architectures and show that the proposed method is able to provide advantage over the standard ELBO formulation in most cases. As this task is simple, all models perform rather well and there is no single best method.

  \section{Concluding Remarks} \label{sec:conclusions}
We presented a principled approach to incorporate prior knowledge about the distribution of predicted scores in Bayesian neural network training. Technically, we augmented the data with a summary observation $s_0$ that captured the prior knowledge. One way to think about the summary is to interpret it as a `pseudo-observation'; pseudo-observations are often used to interpret common priors \citep{gelman2014bayesian}. In order to incorporate all prior knowledge into the model, we apply Bayes rule, and multiply the weight prior with a likelihood term for the summary statistic. Consequently, this yields a well-defined ELBO corresponding to proper Bayesian inference. Thorough empirical experiments in computer vision and natural language processing showed that the proposed method improved robustness and calibration of the BNNs. While we only considered MFVI training of the BNNs, the summary likelihood can be easily incorporated in other Bayesian training methods like MCMC or even to deterministic networks (See Appendix \ref{appsec:dnn-sl}).

\subsubsection*{Acknowledgements}
This work was supported by the Academy of Finland (Flagship programme: Finnish Center for Artificial Intelligence FCAI, and grants 336033, 352986) and EU (H2020 grant 101016775 and NextGenerationEU). The authors would also like to thank Çağlar Hızlı and Manuel Haussmann for useful discussions and our reviewers for their insightful comments that helped us to improve our paper.

  \bibliographystyle{apalike}
  \bibliography{99_library.bib}

\begin{thebibliography}{}

\bibitem[Amini et~al., 2020]{amini2020deep}
Amini, A., Schwarting, W., Soleimany, A., and Rus, D. (2020).
\newblock Deep evidential regression.
\newblock {\em Advances in Neural Information Processing Systems},
  33:14927--14937.

\bibitem[Andrews and Mallows, 1974]{andrews1974scale}
Andrews, D.~F. and Mallows, C.~L. (1974).
\newblock {Scale mixtures of normal distributions}.
\newblock {\em Journal of the Royal Statistical Society: Series B
  (Methodological)}, 36(1):99--102.

\bibitem[Bishop, 2006]{bishop2006pattern}
Bishop, C.~M. (2006).
\newblock {\em {Pattern Recognition and Machine Learning}}.
\newblock Springer.

\bibitem[Blundell et~al., 2015]{blundell2015weight}
Blundell, C., Cornebise, J., Kavukcuoglu, K., and Wierstra, D. (2015).
\newblock {Weight uncertainty in neural networks}.
\newblock {\em arXiv preprint arXiv:1505.05424}.

\bibitem[Cui et~al., 2021a]{cui2021improving}
Cui, T., El~Mekkaoui, K., Havulinna, A., Marttinen, P., and Kaski, S. (2021a).
\newblock {Improving Neural Networks for Genotype-Phenotype Prediction Using
  Published Summary Statistics}.
\newblock {\em bioRxiv}.

\bibitem[Cui et~al., 2021b]{cui2021informative}
Cui, T., Havulinna, A., Marttinen, P., and Kaski, S. (2021b).
\newblock {Informative Bayesian Neural Network Priors for Weak Signals}.
\newblock {\em Bayesian Analysis}, 1(1):1--31.

\bibitem[Deng et~al., 2019]{deng2019adaptive}
Deng, W., Zhang, X., Liang, F., and Lin, G. (2019).
\newblock {An adaptive empirical {B}ayesian method for sparse deep learning}.
\newblock In {\em {Advances in Neural Information Processing Systems}}, pages
  5564--5574.

\bibitem[Dusenberry et~al., 2020]{dusenberry2020efficient}
Dusenberry, M.~W., Jerfel, G., Wen, Y., Ma, Y.-a., Snoek, J., Heller, K.,
  Lakshminarayanan, B., and Tran, D. (2020).
\newblock {Efficient and scalable Bayesian neural nets with rank-1 factors}.
\newblock {\em arXiv preprint arXiv:2005.07186}.

\bibitem[Esteva et~al., 2019]{esteva2019guide}
Esteva, A., Robicquet, A., Ramsundar, B., Kuleshov, V., DePristo, M., Chou, K.,
  Cui, C., Corrado, G., Thrun, S., and Dean, J. (2019).
\newblock {{A guide to deep learning in healthcare}}.
\newblock {\em Nature Medicine}, 25(1):24--29.

\bibitem[Flam-Shepherd et~al., 2017]{flam2017mapping}
Flam-Shepherd, D., Requeima, J., and Duvenaud, D. (2017).
\newblock {Mapping {G}aussian process priors to {B}ayesian neural networks}.
\newblock In {\em {NIPS {B}ayesian deep learning workshop}}.

\bibitem[Fortuin, 2021]{fortuin2021priors}
Fortuin, V. (2021).
\newblock {Priors in Bayesian deep learning: A review}.
\newblock {\em arXiv preprint arXiv:2105.06868}.

\bibitem[Gal and Ghahramani, 2016]{gal2016dropout}
Gal, Y. and Ghahramani, Z. (2016).
\newblock {Dropout as a Bayesian approximation: Representing model uncertainty
  in deep learning}.
\newblock In {\em {International Conference on Machine Learning}}, pages
  1050--1059.

\bibitem[Gelman, 2021]{gelman2021stan}
Gelman, A. (2021).
\newblock {Multiple priors for the same parameter - Stan Forum}.
\newblock
  \url{https://discourse.mc-stan.org/t/multiple-priors-for-the-same-parameter/10943/9}.
\newblock [Online; accessed 11-Oct-2022].

\bibitem[Gelman et~al., 2014]{gelman2014bayesian}
Gelman, A., Carlin, J., Stern, H., Dunson, D., Vehtari, A., and Rubin, D.
  (2014).
\newblock {{Bayesian Data Analysis. vol. 2 CRC press}}.
\newblock {\em Boca Raton, FL}.

\bibitem[George and Huerta, 2018]{george2018deep}
George, D. and Huerta, E.~A. (2018).
\newblock {{Deep Learning for real-time gravitational wave detection and
  parameter estimation: Results with Advanced LIGO data}}.
\newblock {\em Physics Letters B}, 778:64--70.

\bibitem[Ghosh et~al., 2018]{ghosh2018structured}
Ghosh, S., Yao, J., and Doshi-Velez, F. (2018).
\newblock {Structured variational learning of {B}ayesian neural networks with
  horseshoe priors}.
\newblock In {\em {International Conference on Machine Learning}}, pages
  1739--1748.

\bibitem[Graves, 2011]{graves2011practical}
Graves, A. (2011).
\newblock {Practical variational inference for neural networks}.
\newblock In {\em {Advances in Neural Information Processing Systems}}, pages
  2348--2356.

\bibitem[Guo et~al., 2017]{guo2017calibration}
Guo, C., Pleiss, G., Sun, Y., and Weinberger, K.~Q. (2017).
\newblock {On calibration of modern neural networks}.
\newblock In {\em {International Conference on Machine Learning}}, pages
  1321--1330. PMLR.

\bibitem[Hafner et~al., 2018]{hafner2018noise}
Hafner, D., Tran, D., Lillicrap, T., Irpan, A., and Davidson, J. (2018).
\newblock {Noise contrastive priors for functional uncertainty}.
\newblock {\em arXiv preprint arXiv:1807.09289}.

\bibitem[Hendrycks and Dietterich, 2019]{hendrycks2019robustness}
Hendrycks, D. and Dietterich, T. (2019).
\newblock Benchmarking neural network robustness to common corruptions and
  perturbations.
\newblock {\em International Conference on Learning Representations}.

\bibitem[Kingma and Welling, 2013]{kingma2013auto}
Kingma, D.~P. and Welling, M. (2013).
\newblock {Auto-encoding variational Bayes}.
\newblock {\em arXiv preprint arXiv:1312.6114}.

\bibitem[Lakshminarayanan et~al., 2017]{lakshminarayanan2017simple}
Lakshminarayanan, B., Pritzel, A., and Blundell, C. (2017).
\newblock {Simple and scalable predictive uncertainty estimation using deep
  ensembles}.
\newblock {\em Advances in Neural Information Processing Systems}, 30.

\bibitem[Louizos et~al., 2017]{louizos2017bayesian}
Louizos, C., Ullrich, K., and Welling, M. (2017).
\newblock {Bayesian compression for deep learning}.
\newblock In {\em {Advances in Neural Information Processing Systems}}, pages
  3288--3298.

\bibitem[MacKay, 1992]{mackay1992practical}
MacKay, D.~J. (1992).
\newblock {A practical {B}ayesian framework for backpropagation networks}.
\newblock {\em Neural Computation}, 4(3):448--472.

\bibitem[MacKay, 1994]{mackay1994bayesian}
MacKay, D.~J. (1994).
\newblock {Bayesian nonlinear modeling for the prediction competition}.
\newblock {\em ASHRAE transactions}, 100(2):1053--1062.

\bibitem[Malinin and Gales, 2018]{malinin2018predictive}
Malinin, A. and Gales, M. (2018).
\newblock Predictive uncertainty estimation via prior networks.
\newblock {\em Advances in Neural Information Processing Systems}, 31.

\bibitem[Minderer et~al., 2021]{minderer2021revisiting}
Minderer, M., Djolonga, J., Romijnders, R., Hubis, F., Zhai, X., Houlsby, N.,
  Tran, D., and Lucic, M. (2021).
\newblock {Revisiting the calibration of modern neural networks}.
\newblock {\em Advances in Neural Information Processing Systems}, 34.

\bibitem[Molchanov et~al., 2017]{molchanov2017variational}
Molchanov, D., Ashukha, A., and Vetrov, D. (2017).
\newblock {Variational dropout sparsifies deep neural networks}.
\newblock In {\em International Conference on Machine Learning}, pages
  2498--2507. JMLR. org.

\bibitem[Mu and Gilmer, 2019]{mu2019mnist}
Mu, N. and Gilmer, J. (2019).
\newblock {MNIST-C: A robustness benchmark for computer vision}.
\newblock {\em arXiv preprint arXiv:1906.02337}.

\bibitem[Murphy, 2012]{murphy2012machine}
Murphy, K.~P. (2012).
\newblock {\em {Machine Learning: A Probabilistic Perspective}}.
\newblock MIT press.

\bibitem[Nalisnick et~al., 2019]{nalisnick2019dropout}
Nalisnick, E., Hernandez-Lobato, J.~M., and Smyth, P. (2019).
\newblock {Dropout as a structured shrinkage prior}.
\newblock In {\em {International Conference on Machine Learning}}, pages
  4712--4722.

\bibitem[Neal, 1992]{neal1992bayesian}
Neal, R.~M. (1992).
\newblock {Bayesian mixture modeling}.
\newblock In {\em {Maximum Entropy and Bayesian Methods}}, pages 197--211.
  Springer.

\bibitem[Neal, 2012]{neal2012bayesian}
Neal, R.~M. (2012).
\newblock {\em {Bayesian Learning for Neural Networks}}, volume 118.
\newblock Springer Science \& Business Media.

\bibitem[Neklyudov et~al., 2017]{neklyudov2017structured}
Neklyudov, K., Molchanov, D., Ashukha, A., and Vetrov, D.~P. (2017).
\newblock {Structured Bayesian pruning via log-normal multiplicative noise}.
\newblock In {\em {Advances in Neural Information Processing Systems}}, pages
  6775--6784.

\bibitem[Netzer et~al., 2011]{netzer2011reading}
Netzer, Y., Wang, T., Coates, A., Bissacco, A., Wu, B., and Ng, A.~Y. (2011).
\newblock Reading digits in natural images with unsupervised feature learning.
\newblock {\em NIPS Workshop on Deep Learning and Unsupervised Feature
  Learning}.

\bibitem[Paszke et~al., 2019]{paszke2019pytorch}
Paszke, A., Gross, S., Massa, F., Lerer, A., Bradbury, J., Chanan, G., Killeen,
  T., Lin, Z., Gimelshein, N., Antiga, L., et~al. (2019).
\newblock {Pytorch: An imperative style, high-performance deep learning
  library}.
\newblock {\em Advances in Neural Information Processing Systems}, 32.

\bibitem[Reimers and Gurevych, 2019]{reimers2019sentence}
Reimers, N. and Gurevych, I. (2019).
\newblock Sentence-{BERT}: {S}entence embeddings using siamese {BERT}-networks.
\newblock In {\em Conference on Empirical Methods in Natural Language
  Processing}. ACL.

\bibitem[Rieger et~al., 2020]{rieger2020interpretations}
Rieger, L., Singh, C., Murdoch, W., and Yu, B. (2020).
\newblock {Interpretations are useful: penalizing explanations to align neural
  networks with prior knowledge}.
\newblock In {\em {International Conference on Machine Learning}}, pages
  8116--8126. PMLR.

\bibitem[Ross et~al., 2017]{ross2017right}
Ross, A.~S., Hughes, M.~C., and Doshi-Velez, F. (2017).
\newblock {Right for the right reasons: training differentiable models by
  constraining their explanations}.
\newblock In {\em International Joint Conference on Artificial Intelligence},
  pages 2662--2670.

\bibitem[Seeger, 2004]{seeger2004gaussian}
Seeger, M. (2004).
\newblock {Gaussian processes for machine learning}.
\newblock {\em International journal of neural systems}, 14(02):69--106.

\bibitem[Sensoy et~al., 2018]{sensoy2018evidential}
Sensoy, M., Kaplan, L., and Kandemir, M. (2018).
\newblock Evidential deep learning to quantify classification uncertainty.
\newblock {\em Advances in Neural Information Processing Systems}, 31.

\bibitem[Silver et~al., 2016]{silver2016mastering}
Silver, D., Huang, A., Maddison, C.~J., Guez, A., Sifre, L., Van Den~Driessche,
  G., Schrittwieser, J., Antonoglou, I., Panneershelvam, V., Lanctot, M.,
  et~al. (2016).
\newblock {Mastering the game of Go with deep neural networks and tree search}.
\newblock {\em Nature}, 529(7587):484--489.

\bibitem[Socher et~al., 2013]{socher2013recursive}
Socher, R., Perelygin, A., Wu, J., Chuang, J., Manning, C.~D., Ng, A., and
  Potts, C. (2013).
\newblock Recursive deep models for semantic compositionality over a sentiment
  treebank.
\newblock In {\em Conference on Empirical Methods in Natural Language
  Processing}, pages 1631--1642. ACL.

\bibitem[Sun et~al., 2017]{sun2017learning}
Sun, S., Chen, C., and Carin, L. (2017).
\newblock {Learning structured weight uncertainty in Bayesian neural networks}.
\newblock In {\em {Artificial Intelligence and Statistics}}, pages 1283--1292.

\bibitem[Sun et~al., 2019]{sun2019functional}
Sun, S., Zhang, G., Shi, J., and Grosse, R. (2019).
\newblock {Functional variational {B}ayesian neural networks}.
\newblock {\em arXiv preprint arXiv:1903.05779}.

\bibitem[Swiatkowski et~al., 2020]{swiatkowski2020k}
Swiatkowski, J., Roth, K., Veeling, B.~S., Tran, L., Dillon, J.~V., Mandt, S.,
  Snoek, J., Salimans, T., Jenatton, R., and Nowozin, S. (2020).
\newblock {The k-tied normal distribution: A compact parameterization of
  Gaussian mean field posteriors in Bayesian neural networks}.
\newblock {\em arXiv preprint arXiv:2002.02655}.

\bibitem[Teh, 2010]{teh2010dirichlet}
Teh, Y.~W. (2010).
\newblock {Dirichlet Process.}
\newblock {\em Encyclopedia of machine learning}, 1063:280--287.

\bibitem[Teh et~al., 2006]{teh2006hierarchical}
Teh, Y.~W., Jordan, M.~I., Beal, M.~J., and Blei, D.~M. (2006).
\newblock {Hierarchical Dirichlet processes}.
\newblock {\em Journal of the American Statistical Association},
  101(476):1566--1581.

\bibitem[Tran et~al., 2020]{tran2020all}
Tran, B.-H., Rossi, S., Milios, D., and Filippone, M. (2020).
\newblock {All you need is a good functional prior for Bayesian deep learning}.
\newblock {\em arXiv preprint arXiv:2011.12829}.

\bibitem[Tseng et~al., 2020]{tseng2020fourier}
Tseng, A., Shrikumar, A., and Kundaje, A. (2020).
\newblock {Fourier-transform-based attribution priors improve the
  interpretability and stability of deep learning models for genomics}.
\newblock {\em Advances in Neural Information Processing Systems}, 33.

\bibitem[Wang et~al., 2021]{wang2021rethinking}
Wang, D.-B., Feng, L., and Zhang, M.-L. (2021).
\newblock {Rethinking Calibration of Deep Neural Networks: Do Not Be Afraid of
  Overconfidence}.
\newblock {\em Advances in Neural Information Processing Systems}, 34.

\bibitem[Weinberger et~al., 2020]{weinberger2020learning}
Weinberger, E., Janizek, J., and Lee, S.-I. (2020).
\newblock {Learning deep attribution priors based on prior knowledge}.
\newblock {\em Advances in Neural Information Processing Systems},
  33:14034--14045.

\bibitem[Wenzel et~al., 2020]{wenzel2020good}
Wenzel, F., Roth, K., Veeling, B., Swiatkowski, J., Tran, L., Mandt, S., Snoek,
  J., Salimans, T., Jenatton, R., and Nowozin, S. (2020).
\newblock {How Good is the Bayes Posterior in Deep Neural Networks Really?}
\newblock In {\em {International Conference on Machine Learning}}, pages
  10248--10259. PMLR.

\bibitem[Wilson, 2020]{wilson2020case}
Wilson, A.~G. (2020).
\newblock {The case for Bayesian deep learning}.
\newblock {\em arXiv preprint arXiv:2001.10995}.

\bibitem[Wilson and Izmailov, 2020]{wilson2020bayesian}
Wilson, A.~G. and Izmailov, P. (2020).
\newblock {Bayesian deep learning and a probabilistic perspective of
  generalization}.
\newblock {\em arXiv preprint arXiv:2002.08791}.

\bibitem[Xiao et~al., 2017]{xiao2017fashion}
Xiao, H., Rasul, K., and Vollgraf, R. (2017).
\newblock {Fashion-{MNIST}: a Novel Image Dataset for Benchmarking Machine
  Learning Algorithms}.
\newblock {\em arXiv preprint arXiv:1708.07747}.

\bibitem[Yang et~al., 2020]{yang2020incorporating}
Yang, W., Lorch, L., Graule, M., Lakkaraju, H., and Doshi-Velez, F. (2020).
\newblock {Incorporating interpretable output constraints in Bayesian neural
  networks}.
\newblock {\em Advances in Neural Information Processing Systems},
  33:12721--12731.

\end{thebibliography}

  \onecolumn
\appendix

\aistatstitle{Incorporating functional summary information in Bayesian neural networks using a Dirichlet process likelihood approach}

\section{Effect of concentration parameter $\alpha$ in DP} \label{appsec:DP-concentration}

\begin{figure}[!h]
    \centering
    \begin{subfigure}[b]{\textwidth}
        \centering
        \includegraphics[width=\textwidth]{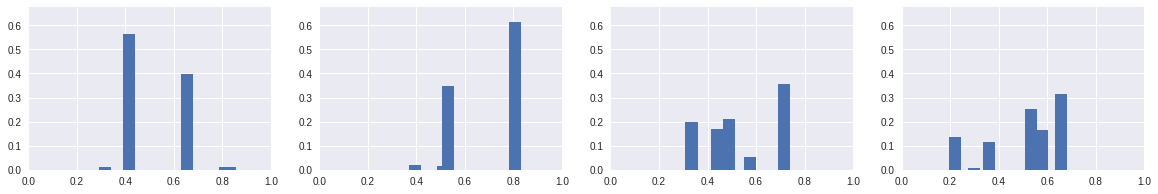}
        \caption{Samples from DP with $\alpha = 1$}
        \label{appfig:dp-a001}
    \end{subfigure}
    \begin{subfigure}[b]{\textwidth}
        \centering
        \includegraphics[width=\textwidth]{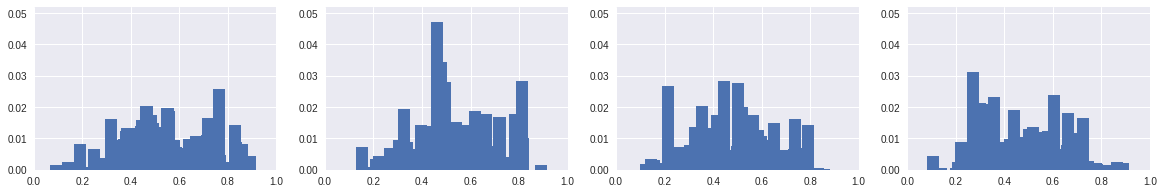}
        \caption{Samples from DP with $\alpha = 100$}
        \label{appfig:dp-a100}
    \end{subfigure}
    \caption{Samples from $\DP(\text{Beta}(5, 5); \alpha)$ for different values of $\alpha$. The concentration parameter $\alpha$ governs how closely the samples from $\DP$ will be to the base measure. In top row, we give the results of using $\alpha = 1$ with a base measure of $\text{Beta}(5,5)$. The samples are widely different, has high variance between them, and individual samples are not close to the base measure used. When we increase the concentration parameter to $\alpha = 100$, the samples drawn from the DP are close to base measure.}
    \label{appfig:dp-alpha}
\end{figure}

As introduced in Sec\ref{sub:dp} and later used in Sec \ref{sub:selbo}, we use Dirichlet
Process to model the summary statistic information $\sobs$. The concentration parameter,
$\alpha$, play a key role in defining the DP and estimating the likelihood loss for each
set of summary samples predicted by neural network. In Fig. \ref{appfig:dp-alpha}, we
show samples drawn from a DP with base measure $\text{Beta}(5, 5)$ for different
values of $\alpha$.

\vfill

\section{Derivation of parameters for $\sobs$} \label{appsec:derivation-prior-params}
In the case of binary classification problem, we assume that the following prior
knowledge is available for modeling the base distribution for Dirichlet Process.
\begin{enumerate}
    \item Fraction of minority class samples.
            If $n_0$ is the number of majority class samples in the training
            set and $n_1$ is the number of minoty class samples, then the
            fraction of minority class samples is defined as
            \begin{equation}
                \gamma_1 = \frac{n_1}{n_0 + n_1}
            \end{equation}
    \item Expected accuracy. This refers to the expected accuracy of a trained
            model and is defined as
            \begin{equation}
                \mathcal{E}_a = \int \limits_{x=0}^{1/2} (1 - x) f(x) dx +
                                    \int \limits_{x=1/2}^{1} x f(x) dx,
            \end{equation}
            where $f(x)$ is the density function of the predicted scores $x$.
\end{enumerate}
\subsection{Useful results}
    \begin{enumerate}
        \item Let $a, b > 0$ and $0 < u < 1$, then
                \begin{equation}
                    \int u^{a - 1} (1 - u)^{b - 1} du = \frac{u^a}{a} {}_2 F_1(a, 1 - b; a + 1; u) + constant, \label{eqn:beta-hyper}
                \end{equation}
                where ${}_2 F_1(\cdot)$ is the hypergeometric function.
    \end{enumerate}

\subsection{Beta distribution as $\sobs$}
The probability density function is defined as
\begin{align}
    f(x; a, b) = \frac{1}{B(a, b)} x^{a-1} (1-x)^{b-1},
\end{align}
where $a, b > 0$, $x \in (0, 1)$ and $B(a, b) = \frac{\Gamma(a)\Gamma(b)}{\Gamma(a+b)}$
is the Beta function.

The cummulative distribution function is defined as
\begin{equation}
    F(x) = \frac{B(x; a, b)}{B(a, b)},
\end{equation}
where $B(x; a, b) = \int \limits_0^{x} t^{a-1} (1-t)^{b-1}dt$ is the incomplete
Beta function. We can apply (\ref{eqn:beta-hyper}) to express this in terms of
hypergeometric function.

Mean of Beta distributed random variable is defined as
\begin{align}
    \mu = \frac{a}{a + b}.
\end{align}

\subsection{Deriving parameters from prior information}
    Let $\gamma_0 = 1 - \gamma_1$ is the fraction of majority samples in the
    training dataset.

    Assuming $s = \frac{1}{2}$ as the threshold for binary decision making,
    where $s$ is the score predicted by the model, we can see that
    \begin{equation}
        \int \limits_{s=0}^{1/2} f(s) ds = \gamma_0.
    \end{equation}

    Assuming that score is distributed as Beta distribution, we have
    \begin{align}
        F(1/2) &= \gamma_0 \nonumber \\
        \frac{B(1/2; a, b)}{B(a, b)} &= \gamma_0 \nonumber \\
        \frac{1}{a 2^a} {}_2 F_1(a, 1 - b; a + 1; 1/2) &= \gamma_0 B(a, b) \label{eqn:beta-cond1}
    \end{align}
    \emph{Unable to proceed because of lack of closed for expression for ${}_2 F_1(\cdot)$}

    Now, consider the expected accuracy $\mathcal{E}_a$,
    \begin{align}
        \mathcal{E}_a &= \int \limits_{0}^{1/2} (1 - s) f(s) ds 
                            + \int \limits_{s=1/2}^{1} s f(s) ds \nonumber \\
            &= \int \limits_{0}^{1/2} f(s) ds - 
                    \int \limits_{0}^{1/2} s f(s) ds + 
                    \int \limits_{s=1/2}^{1} s f(s) ds \label{eqn:ea-terms}
    \end{align}
    
    Applying the density function, we have
    \begin{align}
        \int s f(s) ds &= \int s \cdot \frac{1}{B(a, b)} s^{a-1} (1-s)^{b-1} ds  \nonumber \\
                &= \frac{1}{B(a, b)} \int s^a (1-s)^{b-1} ds \nonumber \\
                &= \frac{1}{B(a, b)} \frac{s^{a+1}}{a+1} {}_2 F_1(a+1, 1-b; a+2; s)
    \end{align}

    Continuing from (\ref{eqn:ea-terms}),
    \begin{align}
        \mathcal{E}_a &= F(1/2) - \frac{1}{B(a, b)} \left( 
                            \left[ 
                                \frac{s^{a+1}}{a+1} {}_2 F_1(a+1, 1-b; a+2; x)
                            \right]_{s=0}^{s=1/2} -
                            \left[ 
                                \frac{s^{a+1}}{a+1} {}_2 F_1(a+1, 1-b; a+2; x)
                            \right]_{s=1/2}^{s=1}
                        \right) \nonumber \\
            &= \frac{1}{2^a B(a, b)}\left(
                    \frac{1}{a} {}_2 F_1(a, 1-b; a+1; 1/2) 
                    \frac{1}{a+1} \left( 
                        {}_2 F_1(a+1, 1-b; a+2; 1/2) - {}_2 F_1(a+1, 1-b; a+2; 1) 
                    \right)
                \right) \label{eqn:beta-cond2}
    \end{align}

    Ideally, from (\ref{eqn:beta-cond1}) and (\ref{eqn:beta-cond2}), we can find the value of $a$ and $b$ given $\gamma_0$ and $\mathcal{E}_a$.    However, closed form expression for solutions are not available. But, using optimization techniques, we can find the approximate solution for this problem. For our experiments, the objective function for optimization is chosen as the MSE between the target value $(\gamma_0, \mathcal{E}_a)$ and the value $(\hat{\gamma}_0, \hat{\mathcal{E}_a})$ observed for the pair $(\hat{a}, \hat{b})$. We used 'L-BFS-G' optimizer available with sklearn package for solving the optimization problem with $\mathcal{E}_a$ set to $0.95$ - $0.98$.

    In multiclass classification experiments, this approach becomes a difficult problem to solve with more unknown that known variables. In those cases, we chose the base distribution parameters by considering the Dirichlet distribution with parameters proportional to number of samples and dividing this into unequal regions to build the base Dirichlet distribution for DP.

\section{SoftHistogram construction during training} \label{appsec:softhist}
The proposed approach requires to estimate the distribution of predicted scorea over a mini-batch of sample to train the model using summary information. We resort to a differentiable histogram for this.

Let $\mathcal{B}$ denotes the regions which constite the prediction simplex and over which we are intereseted in computing the soft histogram. We identity each region by a center $c_i$ and width $\delta_i$ and ensuring that no two regions overlap. For a predicted score vector $\tilde{y}$, the contribution of it towards each region $i$ is computed using
\begin{align}
    g_i(\tilde{y}) = \text{sigmoid}(\sigma * (\tilde{y} - b_i + \delta_i/2)) - \text{sigmoid}(\sigma * (\tilde{y} - b_i - \delta_i/2)).
\end{align}
Here, $\sigma$ acts as the slope of the sigmoid function and effectively improves the quality of histogram estimation. In our experiments, we used $\sigma = 500$.

Finally, over a minibatch $\mathcal{B}$, the weight for each region $i$ is computed as
\begin{align}
    w_i = \sum \limits_{\tilde{y} = f(\x); \x \in \mathcal{B}} g_i(\tilde{y}).
\end{align}

Other methods such as KDE-style estimation with Gaussian kernal can also be used. But in our experiments, we found that using the above method provided stable training.

\section{Cross validation of prior variance} \label{appsec:cv-sigma0}
\begin{figure}[!h]
    \centering
    \begin{subfigure}[b]{0.40\textwidth}
        \centering
        \includegraphics[width=\textwidth]{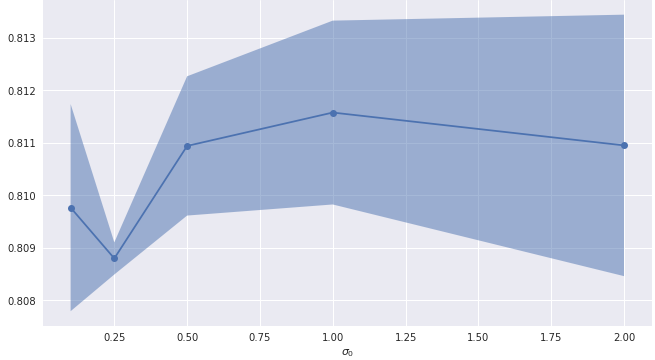}
        \caption{Accuracy vs $\sigma_0$}
        \label{fig:sigma0-cv-acc}
    \end{subfigure}
    \begin{subfigure}[b]{0.40\textwidth}
        \centering
        \includegraphics[width=\textwidth]{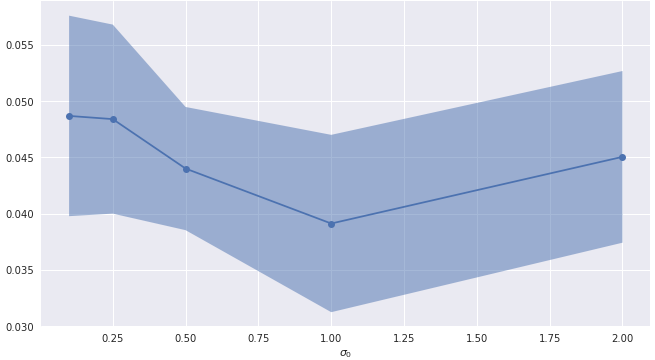}
        \caption{ECE vs $\sigma_0$}
        \label{fig:sigma0-cv-ece}
    \end{subfigure}
    \caption{Results of cross validating prior variance, $\sigma_0$, of neural network parameters. The result is for CIFAR10 with VGG11 architecture.}
    \label{fig:cv-sigma0}
\end{figure}

In this section, we provide results for cross validating the choice of hyperparameters we chose to design the Bayesian Neural Network. The results of crossvalidating prior variance $\sigma_0$ is given in Fig. \ref{fig:cv-sigma0}. Both accuracy and ECE is found to be better at $\sigma_0 = 1$ and we continue to use this value for all our experiments.

\newpage
\section{Additional results for Sentiment Analysis task} \label{appsec:sstbert-additional-results}
\begin{figure}[!h]
    \centering
    \includegraphics[width=\textwidth]{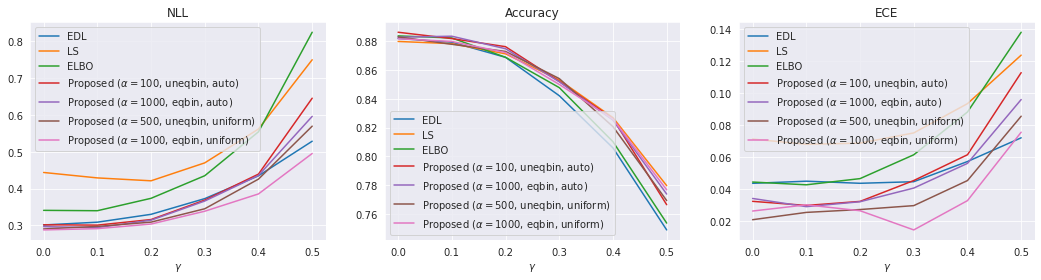}
    \caption{Effect of corruption in sentiment analysis task}
    \label{appfig:sstbert-corruption-comparison}
\end{figure}

In this section, we provide additional results for the sentiment analysis task. In
Table \ref{apptab:sst-indomain-overview}, we provide the in-domain test results and
in Fig. \ref{appfig:sstbert-corruption-comparison} we provide the effect of corruption
in embeddings.

In the labels below, for proposed method, `auto' means base distribution parameters
are computed based on the method in Sec. \ref{appsec:derivation-prior-params}, `uniform'
refers to uniform base distribution, `eqbin' refer to equal bin width strategy and
`uneqbin' refers to unequal bin width strategy discussed in Sec. \ref{ss:sentiment-analysis}
in main text.

\begin{table}[!h]
    \caption{Results on test dataset. Models are trained on SST. The parameters for each
        model are chosen based on best validation NLL.}
    \label{apptab:sst-indomain-overview}
    \centering
    \small
    \begin{tabular}{l c c c c}
        \toprule
        Method & NLL$^{\downarrow}$ & Accuracy$^{\uparrow}$ & AUROC$^{\uparrow}$ & ECE$^{\downarrow}$ \\
        \midrule
            ELBO                                      &          $0.341 \pm 0.024$ &          $0.883 \pm 0.004$ &          $0.952 \pm 0.002$ &          $0.045 \pm 0.009$ \\
            LS                                        &          $0.444 \pm 0.029$ &          $0.880 \pm 0.002$ &          $0.950 \pm 0.001$ &          $0.071 \pm 0.004$ \\
            EDL                                       &          $0.301 \pm 0.001$ &          $0.882 \pm 0.001$ &          $0.954 \pm 0.000$ &          $0.044 \pm 0.004$ \\
            Proposed (auto, eqbin, $\alpha=10^3$)     &          $0.297 \pm 0.010$ &          $0.883 \pm 0.001$ &          $0.953 \pm 0.001$ &          $0.034 \pm 0.004$ \\
            Proposed (uniform, eqbin, $\alpha=10^3$)  & $\mathbf{0.288 \pm 0.002}$ &          $0.881 \pm 0.002$ &          $0.953 \pm 0.000$ &          $0.026 \pm 0.001$ \\
            Proposed (auto, uneqbin, $\alpha=10^2$)   &          $0.302 \pm 0.009$ & $\mathbf{0.886 \pm 0.002}$ & $\mathbf{0.954 \pm 0.001}$ &          $0.033 \pm 0.005$ \\
            Proposed (uniform, uneqbin, $\alpha=500$) &          $0.291 \pm 0.005$ &          $0.883 \pm 0.003$ &          $0.951 \pm 0.002$ & $\mathbf{0.021 \pm 0.005}$ \\
        \bottomrule
    \end{tabular}
\end{table}

\newpage
\section{Results for BinaryMNIST classification task} \label{appsec:bmnist-additional-results}
A binary classification task is constructed from MNIST dataset by sampling only two labels, `3' and `5'. Popular models, LeNet and ConvNet are used as the Bayesian neural network architectures. To study the effect of dataset size, we performed experiments with different sizes - $|\D| = 1000$ and $|\D| = 8000$. We use a summary observation $\sobs$ for the proposed method is constructed using the method described in Appendix \ref{appsec:derivation-prior-params}. Unequal width regions are used to construct the base distribution using $\{0.01, 0.05, 0.10, 0.90, 0.95, 0.99\}$ as boundaries. This unequal width regions help the model to concentrate more of high confidence predictions to match the base distribution while giving less importance to low confidence predictions. The results are provided in Appendix \ref{appsec:bmnist-additional-results} and show that the proposed method is able to provide improved performance in big architectures and large dataset regime.

\begin{table}[!h]
    \caption{Results on clean dataset. Models are trained on MNIST for binary classification. The parameters for each
        model are chosen based on best validation NLL.}
    \label{apptab:bmnist-indomain-overview}
    \centering
    \small
    \begin{tabular}{l c l c c c c}
        \toprule
        & $|\D|$ & Method & NLL$^{\downarrow}$ & Accuracy$^{\uparrow}$ & AUROC$^{\uparrow}$ & ECE$^{\downarrow}$ \\
        \midrule
        \parbox[t]{2mm}{\multirow{8}{*}{\rotatebox[origin=c]{90}{LeNet}}}
            & \parbox[t]{2mm}{\multirow{4}{*}{\rotatebox[origin=c]{90}{$1000$}}}
                & ELBO                      &          $0.013 \pm 0.001$ &          $0.995 \pm 0.001$ &          $1.000 \pm 0.000$ & $\mathbf{0.002 \pm 0.001}$ \\
            &   & LS                        & $\mathbf{0.012 \pm 0.001}$ & $\mathbf{0.995 \pm 0.000}$ & $\mathbf{1.000 \pm 0.000}$ &          $0.003 \pm 0.000$ \\
            &   & EDL                       &          $0.035 \pm 0.002$ &          $0.992 \pm 0.001$ &          $1.000 \pm 0.000$ &          $0.010 \pm 0.002$ \\
            &   & Proposed ($\alpha=50.0$)  &          $0.014 \pm 0.001$ &          $0.994 \pm 0.001$ &          $1.000 \pm 0.000$ &          $0.003 \pm 0.000$ \\
        \cmidrule(l){2-7}
            & \parbox[t]{2mm}{\multirow{4}{*}{\rotatebox[origin=c]{90}{$8000$}}}
                & ELBO                      &          $0.028 \pm 0.003$ &          $0.991 \pm 0.000$ &          $1.000 \pm 0.000$ &          $0.004 \pm 0.000$ \\
            &   & LS                        & $\mathbf{0.019 \pm 0.001}$ & $\mathbf{0.992 \pm 0.001}$ & $\mathbf{1.000 \pm 0.000}$ & $\mathbf{0.003 \pm 0.000}$ \\
            &   & EDL                       &          $0.045 \pm 0.003$ &          $0.989 \pm 0.001$ &          $1.000 \pm 0.000$ &          $0.006 \pm 0.001$ \\
            &   & Proposed ($\alpha=100.0$) &          $0.026 \pm 0.002$ &          $0.992 \pm 0.001$ &          $1.000 \pm 0.000$ &          $0.005 \pm 0.001$ \\
        \midrule
        \parbox[t]{2mm}{\multirow{8}{*}{\rotatebox[origin=c]{90}{ConvNet}}}
            & \parbox[t]{2mm}{\multirow{4}{*}{\rotatebox[origin=c]{90}{$1000$}}}
                & ELBO                      &          $0.039 \pm 0.005$ &          $0.993 \pm 0.001$ &          $1.000 \pm 0.000$ &          $0.024 \pm 0.004$ \\
            &   & LS                        &          $0.040 \pm 0.006$ &          $0.993 \pm 0.001$ & $\mathbf{1.000 \pm 0.000}$ &          $0.024 \pm 0.005$ \\
            &   & EDL                       & $\mathbf{0.030 \pm 0.004}$ & $\mathbf{0.994 \pm 0.000}$ &          $1.000 \pm 0.000$ & $\mathbf{0.012 \pm 0.004}$ \\
            &   & Proposed ($\alpha=10.0$)  &          $0.063 \pm 0.003$ &          $0.990 \pm 0.001$ &          $1.000 \pm 0.000$ &          $0.041 \pm 0.004$ \\
        \cmidrule(l){2-7}
            & \parbox[t]{2mm}{\multirow{4}{*}{\rotatebox[origin=c]{90}{$8000$}}}
                & ELBO                      &          $0.043 \pm 0.005$ &          $0.994 \pm 0.001$ &          $1.000 \pm 0.000$ &          $0.032 \pm 0.005$ \\
            &   & LS                        &          $0.058 \pm 0.018$ &          $0.996 \pm 0.000$ &          $1.000 \pm 0.000$ &          $0.046 \pm 0.016$ \\
            &   & EDL                       &          $0.091 \pm 0.055$ & $\mathbf{0.996 \pm 0.000}$ &          $1.000 \pm 0.000$ &          $0.061 \pm 0.039$ \\
            &   & Proposed ($\alpha=50.0$)  & $\mathbf{0.029 \pm 0.002}$ &          $0.995 \pm 0.001$ & $\mathbf{1.000 \pm 0.000}$ & $\mathbf{0.020 \pm 0.003}$ \\
        \bottomrule
    \end{tabular}
\end{table}

\textbf{In-domain prediction.} In Table \ref{apptab:bmnist-indomain-overview}, we study the performance in in-domain dataset and OOD datasets. Since MNIST is a well curated dataset, there is very less chance of label confusion (as opposite to the sentiment analysis task above) and hence the base distribution we used also encourages the predictions to be concentrated towards high confidence regions, similar to MFVI and LS. However, the proposed method also encourages the distribution of predicted scores to have a non zero mass in low confidence regions and this clearly seems to improve NLL and ECE in clean dataset. We can attribute this to the fact that proposed method makes the models predict less confident scores for incorrect samples and hence lowering the cost of mistake (NLL) and on the course, improving calibration (ECE).

\begin{figure}[t]
    \centering
    \begin{subfigure}[b]{\textwidth}
        \centering
        \includegraphics[width=0.49\textwidth]{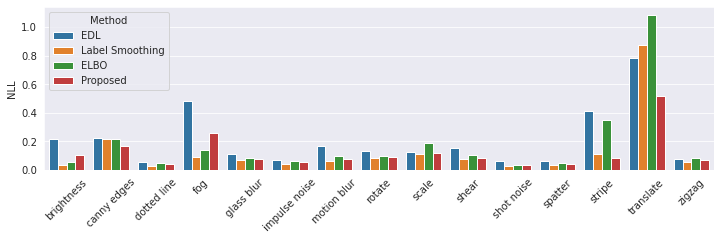}%
        \includegraphics[width=0.49\textwidth]{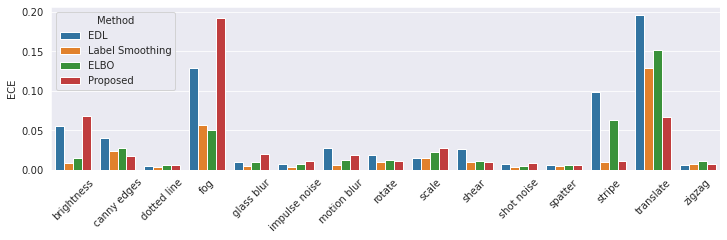}
        \caption{LeNet, $|\mathcal{D}| = 1000$.}
    \end{subfigure}
    \begin{subfigure}[b]{\textwidth}
        \centering
        \includegraphics[width=0.49\textwidth]{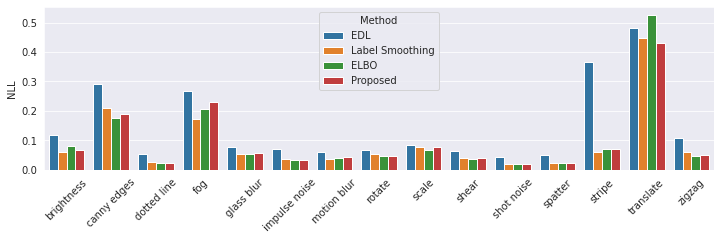}%
        \includegraphics[width=0.49\textwidth]{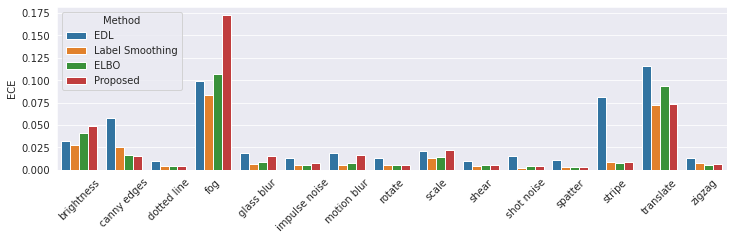}
        \caption{LeNet, $|\mathcal{D}| = 8000$.}
    \end{subfigure}
    \begin{subfigure}[b]{\textwidth}
        \centering
        \includegraphics[width=0.49\textwidth]{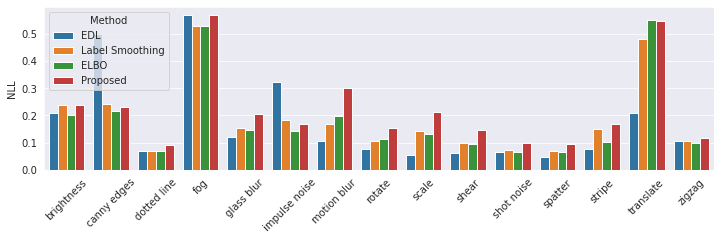}%
        \includegraphics[width=0.49\textwidth]{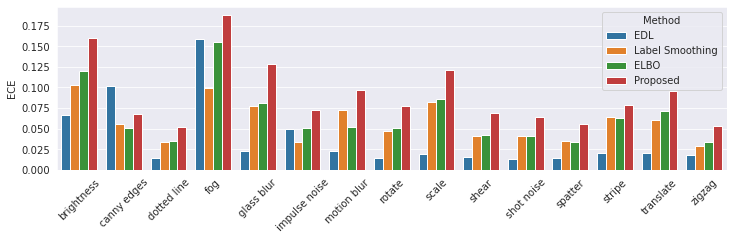}
        \caption{ConvNet, $|\mathcal{D}| = 1000$.}
    \end{subfigure}
    \begin{subfigure}[b]{\textwidth}
        \centering
        \includegraphics[width=0.49\textwidth]{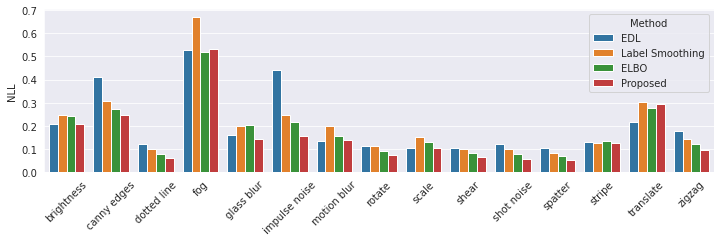}%
        \includegraphics[width=0.49\textwidth]{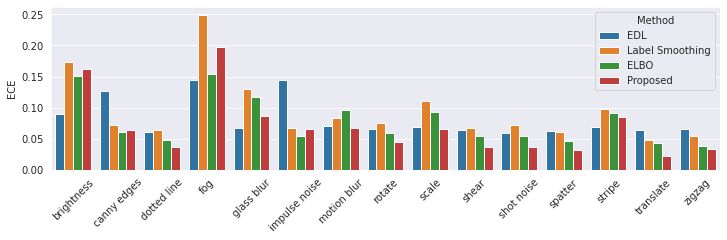}
        \caption{ConvNet, $|\mathcal{D}| = 8000$.}
    \end{subfigure}
    \caption{Comparison of NLL and ECE on different MNIST corruptions. Binary classifier is trained clean MNIST data and tested with various corruptions from MNISTC dataset.}
    \label{appfig:bmnist-comparison}
\end{figure}

\textbf{Corrupted test data.} An important aspect of the proposed method is that it can better represent the uncertainty about predictions by regularizing the posterior score distribution. To test this claim, we provide results on corrupted dataset. We use the MNIST-C \citep{mu2019mnist} dataset which has $15$ corruptions, and we measure the calibration error on the corrupted data. The results in Fig. \ref{appfig:bmnist-comparison} show that in most cases the proposed method is able to perform better than baseline methods.

\begin{table}[!h]
    \caption{Predictive entropy results on OOD setting. Models are trained on clean MNIST dataset and tested on FashionMNIST. The parameters for each model is chosen based on best validation NLL. $\Delta$ denotes the difference between OOD predictive entropy and in-domain predictive entropy.}
    \label{apptab:bmnist-ood-overview}
    \centering
    \small
    \begin{tabular}{l c l c  c c}
        \toprule
        & $|\D|$ & Method & In-domain$^{\downarrow}$ & OOD$^{\uparrow}$ & $\Delta$$^{\uparrow}$ \\
        \midrule
        \parbox[t]{2mm}{\multirow{8}{*}{\rotatebox[origin=c]{90}{LeNet}}}
            & \parbox[t]{2mm}{\multirow{4}{*}{\rotatebox[origin=c]{90}{$1000$}}}
                & ELBO                        & $0.018 \pm 0.001$& $0.317 \pm 0.023$&          $0.300 \pm 0.022$ \\
            &   & LS                          & $0.016 \pm 0.001$& $0.362 \pm 0.011$&          $0.346 \pm 0.011$ \\
            &   & EDL                         & $0.065 \pm 0.004$& $0.191 \pm 0.010$&          $0.126 \pm 0.009$ \\
            &   & Proposed ($\alpha = 50.0$)  & $0.033 \pm 0.001$& $0.488 \pm 0.022$& $\mathbf{0.456 \pm 0.023}$ \\
        \cmidrule(l){2-6}
            & \parbox[t]{2mm}{\multirow{4}{*}{\rotatebox[origin=c]{90}{$8000$}}}
                & ELBO                        & $0.013 \pm 0.001$& $0.433 \pm 0.025$&          $0.421 \pm 0.024$ \\
            &   & LS                          & $0.011 \pm 0.001$& $0.405 \pm 0.017$&          $0.394 \pm 0.016$ \\
            &   & EDL                         & $0.075 \pm 0.006$& $0.276 \pm 0.028$&          $0.201 \pm 0.024$ \\
            &   & Proposed ($\alpha = 100.0$) & $0.017 \pm 0.001$& $0.486 \pm 0.011$& $\mathbf{0.469 \pm 0.011}$ \\
        \midrule
        \parbox[t]{2mm}{\multirow{8}{*}{\rotatebox[origin=c]{90}{ConvNet}}}
            & \parbox[t]{2mm}{\multirow{4}{*}{\rotatebox[origin=c]{90}{$1000$}}}
                & ELBO                        & $0.098 \pm 0.015$& $0.523 \pm 0.015$&          $0.426 \pm 0.009$ \\
            &   & LS                          & $0.100 \pm 0.020$& $0.544 \pm 0.021$&          $0.444 \pm 0.012$ \\
            &   & EDL                         & $0.074 \pm 0.016$& $0.353 \pm 0.028$&          $0.279 \pm 0.031$ \\
            &   & Proposed ($\alpha = 10.0$)  & $0.151 \pm 0.012$& $0.598 \pm 0.012$& $\mathbf{0.446 \pm 0.012}$ \\
        \cmidrule(l){2-6}
            & \parbox[t]{2mm}{\multirow{4}{*}{\rotatebox[origin=c]{90}{$8000$}}}
                & ELBO                        & $0.115 \pm 0.013$& $0.602 \pm 0.013$&          $0.487 \pm 0.010$ \\
            &   & LS                          & $0.156 \pm 0.043$& $0.568 \pm 0.021$&          $0.412 \pm 0.036$ \\
            &   & EDL                         & $0.143 \pm 0.047$& $0.384 \pm 0.056$&          $0.241 \pm 0.018$ \\
            &   & Proposed ($\alpha = 50.0$)  & $0.081 \pm 0.009$& $0.568 \pm 0.011$& $\mathbf{0.487 \pm 0.019}$ \\
        \bottomrule
    \end{tabular}
\end{table}
\textbf{Detection of out-of-distribution samples.} For OOD experiment, we used FashionMNIST \citep{xiao2017fashion} as the test dataset and predictive entropy is measured as a proxy for prediction uncertainty. $\Delta_{\text{OOD}}$ is defined as the difference between OOD and in-domain predictive entropies and higher $\Delta_{\text{OOD}}$ is desirable. From Table \ref{apptab:bmnist-ood-overview}, we can see that the proposed method is able to provide high OOD predictive entropy while maintaining low in-domain predictive entropy.

\clearpage
\section{Additional results for CIFAR10 multiclass classification task} \label{appsec:cifar10-additional-results}
\begin{table}[!h]
    \caption{Results on clean test dataset. Models are trained on CIFAR10. The parameters for each model are chosen based on best validation NLL.}
    \label{apptab:cifar10-indomain-overview}
    \centering
    \small
    \begin{tabular}{l c c c c}
        \toprule
        Method & NLL$^{\downarrow}$ & Accuracy$^{\uparrow}$ & AUROC$^{\uparrow}$ & ECE$^{\downarrow}$ \\
        \midrule
            ELBO                     &          $0.762 \pm 0.011$ &          $0.820 \pm 0.001$ &          $0.978 \pm 0.000$ &          $0.102 \pm 0.002$ \\
            LS ($\epsilon = 0.01$)   &          $1.929 \pm 0.019$ &          $0.787 \pm 0.003$ &          $0.961 \pm 0.001$ &          $0.169 \pm 0.003$ \\
            EDL                      &          $0.779 \pm 0.006$ &          $0.815 \pm 0.001$ &          $0.959 \pm 0.001$ & $\mathbf{0.078 \pm 0.002}$ \\
            Proposed ($\alpha=1000$) & $\mathbf{0.681 \pm 0.003}$ & $\mathbf{0.820 \pm 0.001}$ & $\mathbf{0.979 \pm 0.000}$ &          $0.082 \pm 0.002$ \\
        \bottomrule
    \end{tabular}
\end{table}

\begin{table}[!h]
    \caption{Results on corrupted test dataset. Models are trained on CIFAR10. The parameters for each model are chosen based on best validation NLL on testset. CIFAR10-C is used as corrupted dataset. Results are averaged over all $19$ corruptions and $5$ severity levels.}
    \label{apptab:cifar10-corr-overview}
    \centering
    \small
    \begin{tabular}{l c c c c}
        \toprule
        Method & NLL$^{\downarrow}$ & Accuracy$^{\uparrow}$ & AUROC$^{\uparrow}$ & ECE$^{\downarrow}$ \\
        \midrule
            ELBO                     &          $1.336 \pm 0.022$ & $\mathbf{0.700 \pm 0.004}$ & $\mathbf{0.943 \pm 0.002}$ &          $0.177 \pm 0.003$ \\
            LS ($\epsilon = 0.01$)   &          $3.361 \pm 0.087$ &          $0.666 \pm 0.006$ &          $0.913 \pm 0.003$ &          $0.270 \pm 0.006$ \\
            EDL                      &          $1.313 \pm 0.020$ &          $0.677 \pm 0.005$ &          $0.893 \pm 0.003$ & $\mathbf{0.157 \pm 0.003}$ \\
            Proposed ($\alpha=1000$) & $\mathbf{1.232 \pm 0.022}$ & $\mathbf{0.701 \pm 0.004}$ & $\mathbf{0.945 \pm 0.002}$ & $\mathbf{0.155 \pm 0.003}$ \\
        \bottomrule
    \end{tabular}
\end{table}

\begin{table}[!h]
    \caption{Predictive entropy results on OOD setting. Models are trained on clean CIFAR10
        dataset and tested on SVHN. The parameters for each model is chosen
        based on best validation NLL. $\Delta$ denotes the difference between ood 
        predictive entropy and indomain predictive entropy.}
    \label{apptab:cifar10-ood-overview}
    \centering
    \small
    \begin{tabular}{l c c c}
        \toprule
        Method & In-domain$^{\downarrow}$ & OOD$^{\uparrow}$ & $\Delta$$^{\uparrow}$ \\
        \midrule
        ELBO                     & $0.219 \pm 0.017$& $0.771 \pm 0.089$&          $0.552 \pm 0.077$ \\
        LS ($\epsilon = 0.01$)   & $0.106 \pm 0.003$& $0.240 \pm 0.029$&          $0.134 \pm 0.031$ \\
        EDL                      & $0.481 \pm 0.005$& $1.282 \pm 0.084$& $\mathbf{0.801 \pm 0.083}$ \\
        Proposed ($\alpha=1000$) & $0.268 \pm 0.013$& $0.806 \pm 0.035$&          $0.537 \pm 0.027$ \\
        \bottomrule
    \end{tabular}
\end{table}

\begin{figure}[!h]
    \centering
    \begin{subfigure}[b]{\textwidth}
        \centering
        \includegraphics[width=0.75\textwidth]{figs/cifar10c-vgg11-ece.png}
        \caption{Expected Calibration Error}
        \label{fig:cifar10c-ece}
    \end{subfigure}
    \begin{subfigure}[b]{\textwidth}
        \centering
        \includegraphics[width=0.75\textwidth]{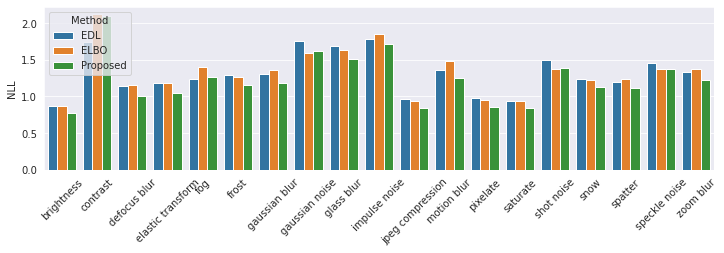}
        \caption{Negative Log Likelihood}
        \label{fig:cifar10c-nll}
    \end{subfigure}
    \begin{subfigure}[b]{\textwidth}
        \centering
        \includegraphics[width=0.75\textwidth]{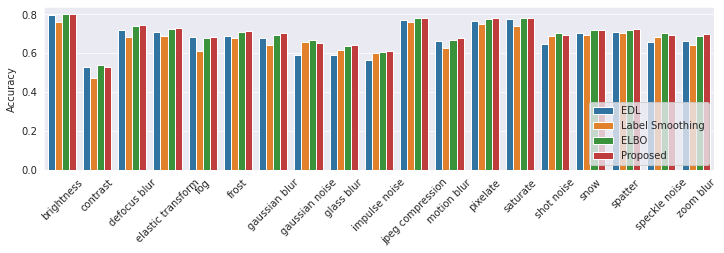}
        \caption{Accuracy}
        \label{fig:cifar10c-acc}
    \end{subfigure}
    \caption{Comparison of performance on corrupt dataset from CIFAR10-C.}
    \label{fig:cifar10c-vgg11-comparison}
\end{figure}

\newpage
\section{Additional results for imbalanced classification problem} \label{appsec:cifar10im-additional-results}
\begin{table}[!h]
    \caption{Results on clean test dataset. Models are trained on Imbalanced CIFAR10. The parameters for each model are chosen based on best validation NLL.}
    \label{apptab:cifar10im-indomain-overview}
    \centering
    \small
    \begin{tabular}{l c c c c}
        \toprule
        Method & NLL$^{\downarrow}$ & F1 Score$^{\uparrow}$ & AUROC$^{\uparrow}$ & ECE$^{\downarrow}$ \\
        \midrule
        ELBO                    &          $1.158 \pm 0.026$ & $\mathbf{0.849 \pm 0.002}$ &          $0.901 \pm 0.003$ &          $0.116 \pm 0.002$ \\
        EDL                     & $\mathbf{0.703 \pm 0.009}$ &          $0.824 \pm 0.003$ &          $0.862 \pm 0.002$ & $\mathbf{0.058 \pm 0.002}$ \\
        Proposed ($\alpha=500$) &          $0.960 \pm 0.022$ & $\mathbf{0.847 \pm 0.001}$ & $\mathbf{0.908 \pm 0.001}$ &          $0.101 \pm 0.003$ \\
        \bottomrule
    \end{tabular}
\end{table}

\begin{table}[!h]
    \caption{Results on corrupted test dataset. Models are trained on Imbalanced CIFAR10. The parameters for each model are chosen based on best validation NLL on test data.}
    \label{apptab:cifar10im-corr-overview}
    \centering
    \small
    \begin{tabular}{l c c c c}
        \toprule
        Method & NLL$^{\downarrow}$ & F1 Score$^{\uparrow}$ & AUROC$^{\uparrow}$ & ECE$^{\downarrow}$ \\
        \midrule
        ELBO                    &          $6.354 \pm 0.117$ &          $0.331 \pm 0.012$ &          $0.421 \pm 0.002$ &          $0.620 \pm 0.010$ \\
        EDL                     & $\mathbf{2.840 \pm 0.034}$ &          $0.314 \pm 0.012$ &          $0.355 \pm 0.002$ & $\mathbf{0.495 \pm 0.007}$ \\
        Proposed ($\alpha=500$) &          $3.564 \pm 0.070$ & $\mathbf{0.400 \pm 0.011}$ & $\mathbf{0.421 \pm 0.003}$ &          $0.525 \pm 0.008$ \\
        \bottomrule
    \end{tabular}
\end{table}

\begin{table}[!h]
    \caption{Predictive entropy results on OOD setting. Models are trained on clean
        imbanalced CIFAR10 dataset and tested on SVHN. The parameters for each model
        are chosen based on best validation NLL. $\Delta$ denotes the difference
        between OOD predictive entropy and indomain predictive entropy.}
    \label{apptab:cifar10im-ood-overview}
    \centering
    \small
    \begin{tabular}{l c c c}
        \toprule
        Method & In-domain$^{\downarrow}$ & OOD$^{\uparrow}$ & $\Delta$$^{\uparrow}$ \\
        \midrule
        ELBO                    & $0.503 \pm 0.020$& $1.461 \pm 0.105$& $\mathbf{0.959 \pm 0.106}$ \\
        EDL                     & $0.099 \pm 0.006$& $0.313 \pm 0.028$&          $0.213 \pm 0.025$ \\
        Proposed ($\alpha=500$) & $0.152 \pm 0.028$& $0.517 \pm 0.075$&          $0.365 \pm 0.051$ \\
        \bottomrule
    \end{tabular}
\end{table}

\begin{figure}[!h]
    \centering
    \begin{subfigure}[b]{\textwidth}
        \centering
        \includegraphics[width=0.65\textwidth]{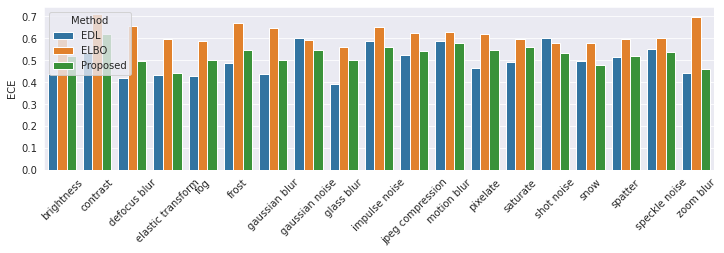}
        \caption{Expected Calibration Error}
        \label{fig:cifar10imc-ece}
    \end{subfigure}
    \begin{subfigure}[b]{\textwidth}
        \centering
        \includegraphics[width=0.65\textwidth]{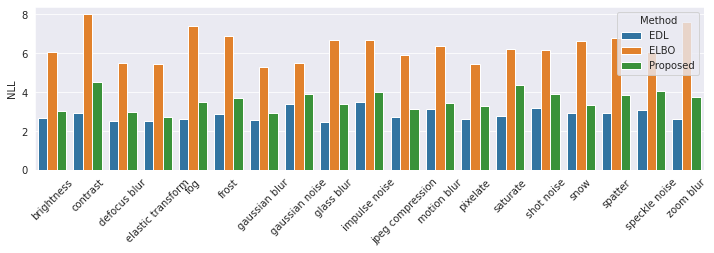}
        \caption{Negative Log Likelihood}
        \label{fig:cifar10imc-nll}
    \end{subfigure}
    \begin{subfigure}[b]{\textwidth}
        \centering
        \includegraphics[width=0.65\textwidth]{figs/cifar10imc-vgg11-f1.png}
        \caption{F1 Score}
        \label{fig:cifar10imc-f1}
    \end{subfigure}
    \caption{Comparison of performance on corrupt dataset from Imbalanced CIFAR10-C.}
    \label{fig:cifar10imc-vgg11-comparison}
\end{figure}

\newpage
\vspace{10cm}
\section{Training deterministic neural networks with Summary Likelihood} \label{appsec:dnn-sl}
Even though in the main section, we focussed on incorporating prior knowledge to training Bayesian Neural Networks, our goal is to introduce a novel \emph{model} to incorporate informative prior information into NNs that is widely applicable without restrictions on the \emph{inference method} (we used standard VI in main section). To demonstrate this flexibility, we provide additional results of using \emph{SGD + momentum} for NN and NN+SL in Tables \ref{apptab:cifar10-indomain-overview-nn} and \ref{apptab:cifar10-ood-overview-nn} (BNNs in main section were trained $5000$ steps, no DA). We observe that data augmentation can improve accuracy but importantly also that the SL consistently improves the accuracy, calibration error and OOD detection. We believe SL can be beneficial with other better inference methods of NNs, both Bayesian and non-Bayesian, such as the deep ensembles. 

\begin{table}[!h]
    \caption{Results on VGG11 trained on CIFAR10. Evaluation is performed on clean dataset. Numbers on braces indicates the number of training steps. SL - Proposed approach. DA - Data augmentation.}
    \label{apptab:cifar10-indomain-overview-nn}
    \centering
    \small
    \begin{tabular}{l c c c c}
        \toprule
        Method & NLL$^{\downarrow}$ & Accuracy$^{\uparrow}$ & AUROC$^{\uparrow}$ & ECE$^{\downarrow}$ \\
        \midrule
        NN ($5k$)                   &          $1.006 \pm 0.003$ & $\mathbf{0.808 \pm 0.001}$ & $\mathbf{0.977 \pm 0.000}$ &          $0.138 \pm 0.001$ \\
        NN + SL ($5k$)       & $\mathbf{0.803 \pm 0.001}$ &          $0.799 \pm 0.001$ &          $0.976 \pm 0.000$ & $\mathbf{0.121 \pm 0.000}$ \\
        NN ($30k$)                  &          $0.948 \pm 0.003$ &          $0.810 \pm 0.001$ &          $0.977 \pm 0.000$ &          $0.133 \pm 0.001$ \\
        NN + SL ($30k$)      & $\mathbf{0.772 \pm 0.005}$ & $\mathbf{0.817 \pm 0.000}$ & $\mathbf{0.980 \pm 0.000}$ & $\mathbf{0.115 \pm 0.001}$ \\
        \midrule
        NN + DA ($5k$)              & $\mathbf{0.498 \pm 0.004}$ & $\mathbf{0.843 \pm 0.001}$ & $\mathbf{0.986 \pm 0.000}$ &          $0.054 \pm 0.001$ \\
        NN + SL + DA ($5k$) &          $0.501 \pm 0.005$ &          $0.834 \pm 0.002$ &          $0.985 \pm 0.000$ & $\mathbf{0.041 \pm 0.001}$ \\
        NN + DA ($30k$)            &          $0.566 \pm 0.004$ & $\mathbf{0.887 \pm 0.001}$ &          $0.991 \pm 0.000$ &          $0.080 \pm 0.001$ \\
        NN + SL + DA ($30k$) & $\mathbf{0.449 \pm 0.002}$ &          $0.886 \pm 0.001$ & $\mathbf{0.992 \pm 0.000}$ & $\mathbf{0.067 \pm 0.000}$ \\
        \bottomrule
    \end{tabular}
\end{table}

\begin{table}[!h]
    \caption{Predictive entropy results on OOD setting. VGG11 model is trained on clean CIFAR10 dataset and tested on SVHN. $\Delta$ denotes the difference between OOD predictive entropy and indomain predictive entropy.}
    \label{apptab:cifar10-ood-overview-nn}
    \centering
    \small
    \begin{tabular}{l c c c}
        \toprule
        Method & In-domain$^{\downarrow}$ & OOD$^{\uparrow}$ & $\Delta$$^{\uparrow}$ \\
        \midrule
        NN ($5k$)               & $0.140 \pm 0.001$& $0.381 \pm 0.010$&          $0.241 \pm 0.010$ \\
        NN + SL ($5k$)   & $0.222 \pm 0.004$& $0.585 \pm 0.045$& $\mathbf{0.363 \pm 0.044}$ \\
        NN ($30k$)            & $0.150 \pm 0.000$& $0.384 \pm 0.011$&          $0.234 \pm 0.012$ \\
        NN + SL ($30k$)  & $0.186 \pm 0.004$& $0.475 \pm 0.021$& $\mathbf{0.290 \pm 0.020}$ \\
        \midrule
        NN + DA ($5k$)         & $0.315 \pm 0.005$& $1.010 \pm 0.038$& $\mathbf{0.695 \pm 0.036}$ \\
        NN + SL + DA ($5k$) & $0.377 \pm 0.008$& $1.065 \pm 0.025$&          $0.689 \pm 0.029$ \\
        NN + SL ($30k$)           & $0.093 \pm 0.001$& $0.365 \pm 0.005$&          $0.272 \pm 0.005$ \\
        NN + SL + DA ($30k$) & $0.135 \pm 0.001$& $0.531 \pm 0.026$& $\mathbf{0.395 \pm 0.026}$ \\
        \bottomrule
    \end{tabular}
\end{table}

\end{document}